\documentclass[10pt,twocolumn,letterpaper]{article}

\usepackage{cvpr}
\usepackage{times}
\usepackage{epsfig}
\usepackage{graphicx}
\usepackage{amsmath}
\usepackage{amssymb}

\usepackage[normalem]{ulem}

\usepackage[table,xcdraw]{xcolor} 

\usepackage{siunitx}
\definecolor{darkred}{rgb}{0.55, 0.0, 0.0}
\definecolor{lavenderindigo}{rgb}{0.58, 0.34, 0.92}
\definecolor{orange}{rgb}{1.0, 0.5, 0.0}

\newcommand{\blue}[1]{{\color{blue}{#1}}}
\newcommand{\red}[1]{{\color{red}{#1}}}
\newcommand{\darkred}[1]{{\color{darkred}{#1}}}
\newcommand{\purple}[1]{{\color{lavenderindigo}{#1}}}
\newcommand{\green}[1]{{\color{green}{#1}}}
\newcommand{\orange}[1]{{\color{orange}{#1}}}

\newcommand{\tikzcircle}[2][black,fill=black]{\tikz[baseline=-0.5ex]\draw[#1,radius=#2] (0,0) circle ;}%

\usepackage{pifont}
\newcommand{\cmark}{\ding{51}}%
\newcommand{\xmark}{\ding{55}}%
\newcommand{\Xmark}{\ding{53}}%

\usepackage{float} 
\usepackage{booktabs} 

\usepackage{caption} 
\usepackage{subcaption} 
\usepackage{wrapfig}
\usepackage{bm} 
\usepackage{bbm} 
\usepackage{makecell} 
\usepackage{multirow} 

\makeatletter
\@namedef{ver@everyshi.sty}{}
\makeatother
\usepackage{tikz}
\DeclareRobustCommand\full  {\tikz[baseline=-0.6ex]\draw[thick] (0,0)--(0.5,0);}

\DeclareRobustCommand\dashed{\tikz[baseline=-0.6ex]\draw[thick,dashed] (0,0)--(0.54,0);}

\let\originalleft\left
\let\originalright\right
\renewcommand{\left}{\mathopen{}\mathclose\bgroup\originalleft}
\renewcommand{\right}{\aftergroup\egroup\originalright}

\newcommand{\modelI}{SiamCT-DFG}
\newcommand{\modelIatt}{SiamCT-DFG-att}
\newcommand{\modelII}{SiamCT-CA}
\newcommand{\modelIIppm}{SiamCT-CA+PPM}

\newcommand{\bb}{\mathbf{b}}
\newcommand{\bS}{\mathbf{S}}

\newcommand{\bv}{\mathbf{v}}

\newcommand{\bh}{\mathbf{h}}

\newcommand{\bX}{\mathbf{X}}

\newcommand{\bW}{\mathbf{W}}

\newcommand{\bA}{\bm{A}}
\newcommand{\bH}{\bm{H}}
\newcommand{\realnumbers}{\mathbb{R}}

\usepackage[pagebackref=true,breaklinks=true,letterpaper=true,colorlinks,bookmarks=false]{hyperref}

\cvprfinalcopy 

\begin{document}

\title{Siamese Tracking with Lingual Object Constraints}

\author{Maximilian Filtenborg\thanks{During this research Maximilian was an intern at the QUVA lab.}\\
QUVA Lab \\
University of Amsterdam\\
{\tt\small max.filtenborg@gmail.com}
\and
Efstratios Gavves\\
QUVA Lab \\
University of Amsterdam\\
{\tt\small egavves@uva.nl}
\and
Deepak Gupta\\
QUVA Lab\\
University of Amsterdam\\
{\tt\small d.k.gupta@uva.nl}
}

\maketitle

\begin{abstract}

Classically, visual object tracking involves following a target object throughout a given video, and it provides us the motion trajectory of the object.
However, for many practical applications, this output is often insufficient since additional semantic information is required to act on the video material.
Example applications of this are surveillance and target-specific video summarization, where the target needs to be monitored with respect to certain predefined constraints, \emph{e.g.}, `when standing near a yellow car'.
This paper explores, \textit{tracking visual objects subjected to additional lingual constraints}.
Differently from Li \etal \cite{Li_2017_CVPR_Natural_Language}, we impose additional lingual constraints upon tracking, which enables new applications of tracking.
Whereas in their work the goal is to improve and extend upon tracking itself.
To perform benchmarks and experiments, we contribute two datasets: c-MOT16 and c-LaSOT, curated through appending additional constraints to the frames of the original LaSOT \cite{DS:Fan2018LaSOTAH} and MOT16 \cite{DS:milan2016mot16} datasets.
We also experiment with two deep models \modelI{} and \modelII{}, obtained through extending a recent state-of-the-art Siamese tracking method and adding modules inspired from the fields of natural language processing and visual question answering. 
Through experimental results, we show that the proposed model \modelII{} can significantly outperform its counterparts.
Furthermore, our method enables the selective compression of videos, based on the validity of the constraint.

\end{abstract}


\section{Introduction} \label{sec:introduction} 
Inspired by the recent advancements in deep learning and the ability to handle multi-modal inputs (in this case in the form of text and videos), we propose to incorporate lingual constraints in the form of sentences in the tracking domain.
In this study, we investigate how to incorporate these constraints effectively within tracking.
If we are interested in only a certain set of frames where the target is involved with a specific object or action, we would have to watch all the images of a potentially very long track (the sequence of frames the target was tracked).
This can be interpreted as imposing additional constraints on the tracking process, rather than only performing similarity learning to match the target to the ground-truth frame.
This constraint is a lingual specification that the set of frames have to be matched to, such that the tracking sequence can be filtered to these frames.


\begin{figure}
    \centering
    \captionsetup[subfigure]{position=t,justification=centering}
    \begin{subfigure}[t]{0.15\textwidth}
         \centering
         \includegraphics[width=\textwidth]{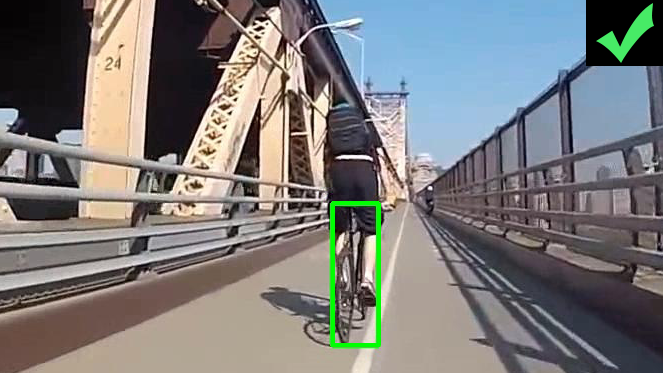}
         \caption*{`with a black backpack'}
         \label{fig:ds_lasot_example_2}
    \end{subfigure}
    \hfill
    \begin{subfigure}[t]{0.15\textwidth}
         \centering
         \includegraphics[width=\textwidth]{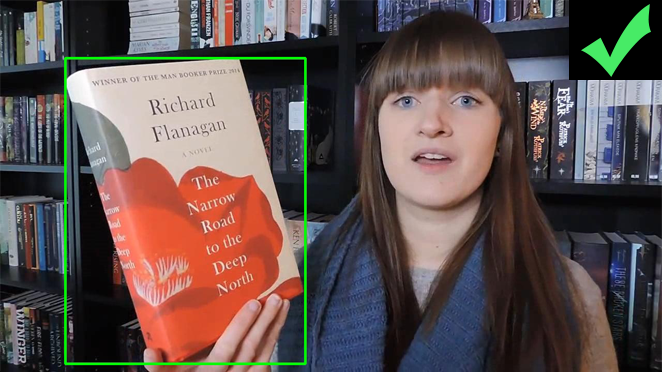}
         \caption*{`close to a person'}
         \label{fig:ds_lasot_example_3}
    \end{subfigure}
    \hfill
    \begin{subfigure}[t]{0.15\textwidth}
     \centering
     \includegraphics[width=\textwidth]{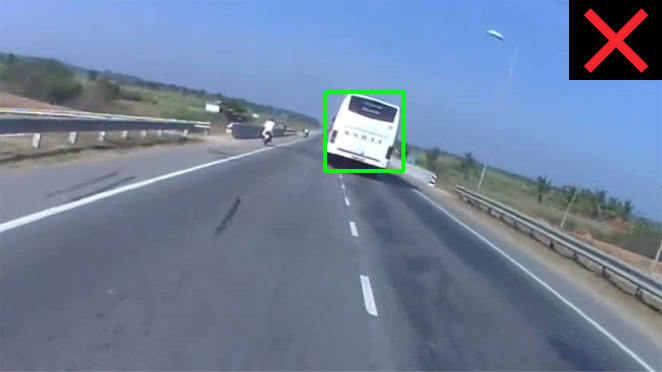}
     \caption*{`adjacent to a car'}
     \label{fig:ds_lasot_example_1}
    \end{subfigure}
    \caption{\textbf{Tracking with visual object constraints.}
    Example images from c-LaSOT dataset where the target object is being tracked and subjected to additional lingual constraints.}
    \label{fig:ds_lasot_examples}
\end{figure}

The problem we address in this work is tracking with a lingual constraint, where the lingual constraint is defined as a sentence provided by the user that describes a certain state or environment of the target.
Since the constraint is not a pre-defined entity, buSurprisingly, \modelIatt{} scores higher on almost all metrics and similarly has `better' Precision-Recall curves and ROC AUC curves.

t in the form of a natural language sentence, the constraint is, in essence, unrestricted in terms of what it can contain or capture.
To evaluate, we extend the popular LaSOT \cite{DS:Fan2018LaSOTAH} and MOT16 \cite{DS:milan2016mot16} datasets with lingual constraints\footnote{Code and data is available at our GitHub: \url{https://github.com/CMFiltenborg/lingually_constrained_tracking}}.

\begin{itemize}
\item We pose a novel research problem of visual tracking of objects subjected to additional lingual constraints. 
\item To evaluate and benchmark said task, we present two new datasets, c-MOT and c-LaSOT, for tracking subject to lingually specified visual object (detection) constraints. 
\item Finally, we experiment with two models, \modelI{} and \modelII{}. For further work we propose \modelII{}, built upon the SiamRPN++ model and specifically extended for tracking with visual object constraints. \modelII{} is inspired by state-of-the-art techniques from the VQA field, and specifically features the MCAN model \cite{VQA:Yu2019DeepMC} and the Pyramid Pooling Module (PPM) \cite{Zhao2017PyramidSP}. Furthermore, we conduct ablation experiments to evaluate the effectiveness of the Pyramid Pooling Module and the importance of attention in the models for the constraint inference. 
\end{itemize}

\section{Related Work} \label{sec:related-work}
Next, we discuss works from tracking, VQA, and natural language tracking.

\textbf{Tracking.} 
For a long time the best trackers were based on discriminative correlation filters \cite{Danelljan2016ECOEC, VOT2017:kristan2017visual}.
Recently the siamese tracking paradigm has gained significant traction \cite{Bertinetto2016FullyConvolutionalSN_siamFCv1, Valmadre_2017_CVPR_siamFCv2, Guo2017LearningDSiam, Li2018HighPV_siamRPN}.
An important improvement upon the siamese tracking architecture came with the introduction of the SiamRPN and SiamRPN++ architectures, which includes the RPN module \cite{Li2018HighPV_siamRPN, Li2018SiamRPNEO_siamRPN++}.
More recent improvements were seen in improving the online tracking capabilities of the deep trackers, by predicting a tracking model online \cite{Bhat2019LearningDM_DIMP}, which fully incorporates the background and foreground information of the target.
Instead of predicting bounding-boxes, the authors of \cite{Bhat2019LearningDM_DIMP, Danelljan2018ATOMAT} introduce methods that predicts the overlap of a particular bounding-box and the target.
In this work, we employ SiamRPN++ \cite{Li2018SiamRPNEO_siamRPN++} as the basis of our tracker.



\textbf{Natural Language Tracking.} 
Li \etal \cite{Li_2017_CVPR_Natural_Language} propose to use a natural language specification to identify the target object.
However, it is important to note that the lingual specification here serves to improve the localization of the target and to track by the natural language specification itself, instead of the bounding box. 
Moreover, this work does not use the lingual specification as an additional constraint. 
Additional discussion related to this work can be found in section \ref{sec:appendix_li_comparison} of the supplementary material.
Gavrilyuk \etal \cite{Gavrilyuk_2018_sentence}, improve upon the Lingual specification network from Li \etal \cite{Li_2017_CVPR_Natural_Language}, to tackle the task of actor and action segmentation in a video. 
Which they are also able to compare to the earlier work of Hu \etal \cite{Hu2016SegmentationFN}, on the task of semantic segmentation based on a natural language query.

\textbf{Visual Question Answering.} 
In the field of Visual Question Answering (VQA) the aim is to answer natural language questions about images. 
To tackle this, the natural language question and the picture have to be reasoned about in tandem.
The domain of VQA inhibits many of the same challenges that we face in the present work, where reasoning about multiple modalities together is the main challenge.
The field of VQA has mostly been dominated by one architectural approach, which is based on the Transformer \cite{Vaswani2017AttentionIA}.

The main building block of the Transformer is the Attention module.
Consequently, it has become a component of all (recent) successful VQA approaches, and various variants have been proposed to improve the attention module for multi-modal reasoning.
Recent works \cite{VQA:Yu2019DeepMC, VQA:Lu2016HierarchicalQC} have shown that co-attention in addition to self-attention is an important step for further improvements.
Yu \etal \cite{VQA:Yu2019DeepMC}, was an influential work, as it beat the SOTA and won the VQA challenge in 2019.
Further progress has been made upon this, by integrating the co-attention and self-attention into one mechanism \cite{VQA:Yu2019MultimodalUA}.
Other work has focused on the important soft-max operation \cite{VQA:Martins2020SparseAS} in the attention module.
Also important is the choice of input image features used, as many works use expensive `region features' \cite{VQA:jiang2020defense, VQA:Farazi2020AccuracyVC}.
Jiang \etal \cite{VQA:jiang2020defense} and Farazi \etal \cite{VQA:Farazi2020AccuracyVC} make the case that the use of region features is unneeded, and that grid features are more easily optimised and can outperform region features.
Lastly, Gokhale \etal \cite{VQA:Gokhale2020VQALOLVQ} shows that the VQA models employed currently are not able to answer logical compositions or negations of natural language questions.

\section{Datasets} \label{sec:datasets}
Since there is no available tracking dataset that contains all the necessary ground truth information for the proposed task, a dataset had to be manually labelled.
We present c-LaSOT and c-MOT16, which have been built through additional annotations of the LaSOT \cite{DS:Fan2018LaSOTAH} and MOT16 \cite{DS:milan2016mot16} datasets respectively.

To determine whether or not the lingual constraint is satisfied, the described constraint (object) has to be `close to' the target object that is being tracked.
We define the constraint to be satisfied when the target object and the constraint object are in close proximity, e.g., if the lingual object constraint is a pencil, and the tracking target is a person, this person has to be holding the pencil or otherwise be near it for the constraint to be positive in the frame.
To track (or find) the target when the constraint is satisfied, the target has to be located in those frames. 
This means that the tracker has to continuously track the target, even while the constraint is unsatisfied. 
Only when the constraint is satisfied, the user also gains this information for these video frames.

\textbf{Constrained LaSOT.}
c-LaSOT features $6$ object classes as constraints, which are present in different videos throughout the original LaSOT \cite{DS:Fan2018LaSOTAH} dataset. 
The LaSOT dataset is a state-of-the-art dataset for tracking, which contains long sequences with $70$ different object classes as tracking targets.
In order to limit the number of annotations required and the complexity of annotating the dataset itself, a selection of $6$ object classes was taken that can act as the constraining object, which were selected based on the availability of the object in the dataset.
Figure \ref{fig:ds_lasot_examples} shows some examples of the Constrained LaSOT (c-LaSOT) dataset, and table \ref{tab:object-classes} depicts a summary of the dataset, further details on the construction of the dataset are specified in the supplementary materials, in section \ref{sec:appendix_ds_lasot}.

\begin{table}
\centering
\caption{\textbf{c-LaSOT summary.} The table depicts the constraining object and the number of annotated sequences they appear in. For more details, we refer to section \ref{sec:ds_lasot_challenges} of the supplementary materials.}
\label{tab:object-classes}
\resizebox{0.45\textwidth}{!}{%
\begin{tabular}{@{}llll@{}}
\toprule
\# & Constraint Object   & \# Sequences & Example                                            \\ \midrule
1  & car      & 14           & `adjacent to a car'                                \\
2  & person   & 69           & \makecell[l]{`close to a person with\\ a black motorcycle helmet'} \\
3  & backpack & 2            & `with a black backpack'                            \\
4  & cat      & 3            & `playing with a grey cat'                          \\
5  & hand     & 10           & `close to a persons hand'                          \\
6  & bottle   & 2            & `near a chardonnay wine bottle'                   \\ \bottomrule
\end{tabular}%
}
\end{table}

\begin{figure}
    \centering
    \captionsetup[subfigure]{position=b,justification=centering}
    \begin{subfigure}[b]{0.23\textwidth}
     \centering
     \includegraphics[width=\textwidth]{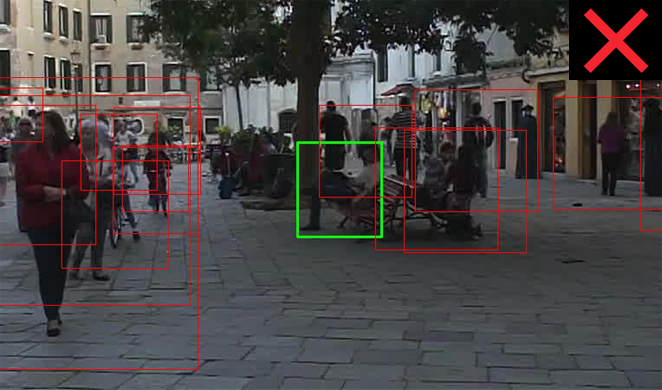}
     \caption*{`a bicycle'}
     \label{fig:ds_mot_example_1}
    \end{subfigure}
    \hfill
    \begin{subfigure}[b]{0.23\textwidth}
         \centering
         \includegraphics[width=\textwidth]{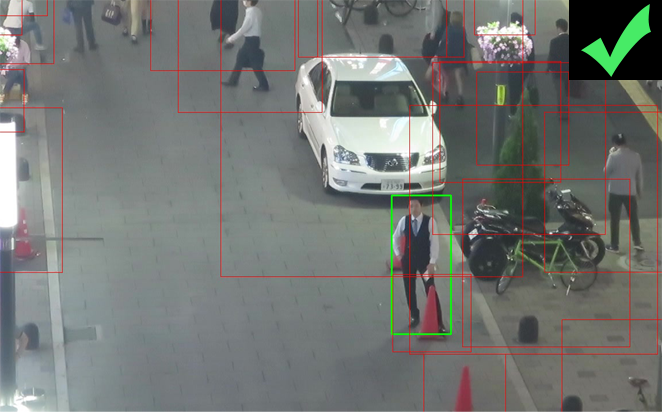}
         \caption*{`a black motorcycle'}
         \label{fig:ds_mot_example_2}
    \end{subfigure}
    \\
    \begin{subfigure}[t]{0.23\textwidth}
         \centering
         \includegraphics[width=\textwidth]{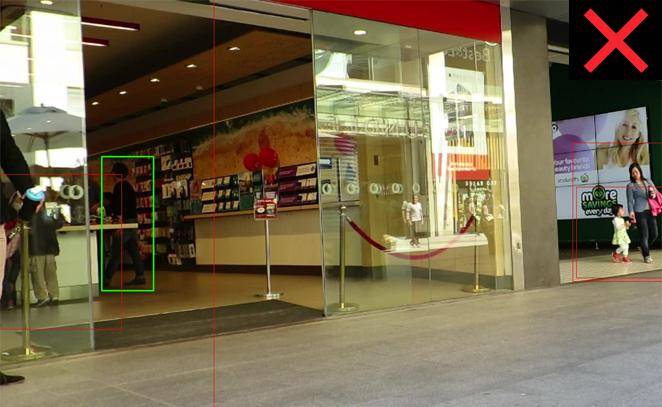}
         \caption*{`a person wearing a denim jacket and khaki pants'}
         \label{fig:ds_mot_example_3}
    \end{subfigure}
    \hfill
    \begin{subfigure}[t]{0.23\textwidth}
         \centering
         \includegraphics[width=\textwidth, height=2.5cm]{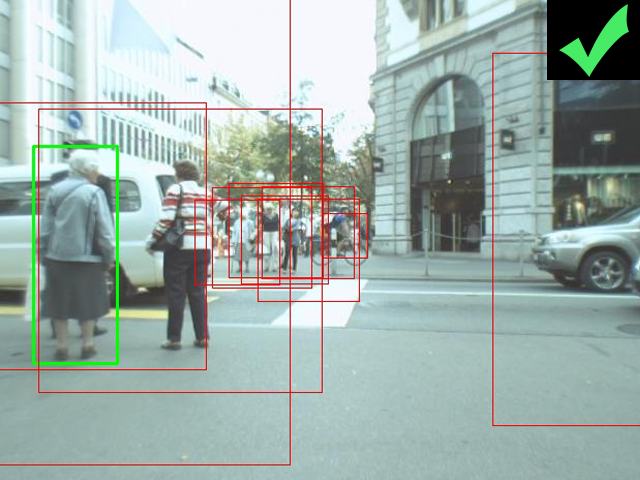}
         \caption*{`near a white van'}
         \label{fig:ds_mot_example_4}
    \end{subfigure}
    \caption{\textbf{c-MOT16 examples.} We display one tracking target (the green bounding box) and all other objects posing as a constraint (the red bounding boxes) for this target in all of its sequences, with an example constraint depicted below. Images are zoomed in for clarity.}
    \label{fig:mot_dataset_examples}
\end{figure}

\textbf{Constrained MOT16.}
\label{sec:mot_dataset}
We introduce c-MOT16, a dataset that is built on top of the MOT16 dataset \cite{DS:milan2016mot16}, and features 14 video sequences, $7$ training videos, and $7$ test videos\footnote{For this research, we can only leverage the training videos, as the annotations of the test videos are not published, and can only be leveraged in the MOT challenge.}, which were filmed with both static and moving cameras.
The MOT data paradigm provides an opportunity to construct a dataset for tracking with visual object constraints, as it has many of the annotations required. 
We leverage the fact that MOT has multiple tracking targets in every frame of its videos, by having one tracking target act as the constraining object, and another tracking target as the real tracking target.
This way, if object $A$ is the tracking target of a sequence, we leverage another tracking target object $B$ as the constraining object, where the natural language constraint of $B$ is a sentence that describes its characteristics.    
Hence, the position and movement of both  $A$ and $B$ are known throughout the video (in the form of bounding-boxes).
With this information, we compute when the constraint is satisfied by determining when $B$ is in proximity of $A$.
Figure \ref{fig:mot_dataset_examples} visually depicts a few examples of the resulting dataset.
For more explicit notation, we introduce the subscript $b$ for an objects bounding box, i.e. $A_b$ is the bounding box of object $A$.
To determine whether or not object $B$ is in proximity of object $A$, we compute the area of overlap, $O$, of the bounding box $B_b$ of object $B$ with the bounding box $A_b$ of object $A$.

After we compute the overlap, $O(B_b, A_b) = \frac{| A_b \cap B_b |}{| B_b |}$. 
The function $Y(A_b, B_b) \in \{0, 1\}$ tells us whether the overlap is past a certain threshold $t = 0.5$, which also functions as the groundtruth for the samples in our dataset, $y_i = Y(A_{b_{i}}, B_{b_{i}})$, where the subscript $i$ indicates the sample index (which is omitted elsewhere for clarity).
Note, that in this research we do not consider the distance of objects $A$ and $B$ in depth relative to each other in the frame.
Table (\ref{tab:mot16-table}) shows a detailed summary of the c-MOT16 dataset.
Further details of the lingual constraints and the creation of the dataset are specified in the supplementary materials, in section \ref{sec:appendix_ds_mot}.

\begin{table}
\centering
\caption{\textbf{c-MOT16 summary.}. This table shows a summary of the dataset split in training, validation, and testing. In the table, `N-videos' denotes the number of videos allocated to that sub-group, `Tracks' is the number of sequences per group.}
\label{tab:mot16-table}
\resizebox{0.45\textwidth}{!}{%
\begin{tabular}{@{}llllll@{}}
\toprule
           & N-Videos & Tracks & Positive & Negative & \% Positive \\ \midrule
Training   & 4        & 9417   & 2.7 M    & 5.8 M    & 31\%        \\
Validation & 1        & 1778   & 118 K    & 116 K    & 50\%        \\
Testing    & 2        & 1648   & 138 K    & 142 K    & 49\%        \\ \midrule
Total      & 7        & 12843  & 3.0 M    & 8.2 M    & 32\%        \\ \bottomrule
\end{tabular}
}%
\end{table}

\section{Tracking with Lingual Object Constraints} \label{sec:methodology} 
In this section, we introduce two architectures to tackle the task of tracking with visual object constraints, in which we first discuss the common components of both, starting with the setup of the deep tracker.

\textbf{Deep Tracker.}
For the tracker, we are inspired by the work of Li \etal \cite{Li2018SiamRPNEO_siamRPN++} and further build upon this model, which features the SiamRPN++ model, one of the recent state-of-the-art deep trackers to perform tracking in videos.
We approach the task of tracking with lingual constraints with a siamese tracking setup, which tackles the task of tracking by posing it as a single shot pattern matching problem.
Since the aim of this work is to extend a tracker to incorporate constraints, and the goal is not to improve the tracking architecture itself, the deep tracker modules are maintained as they are and not optimised\footnote{Hence, the more intricate details of the SiamRPN++ model are considered out of scope for this work.}.

\begin{figure*}
  \centering
  \begin{subfigure}[b]{0.38\textwidth}
     \centering
     \includegraphics[width=1\textwidth]{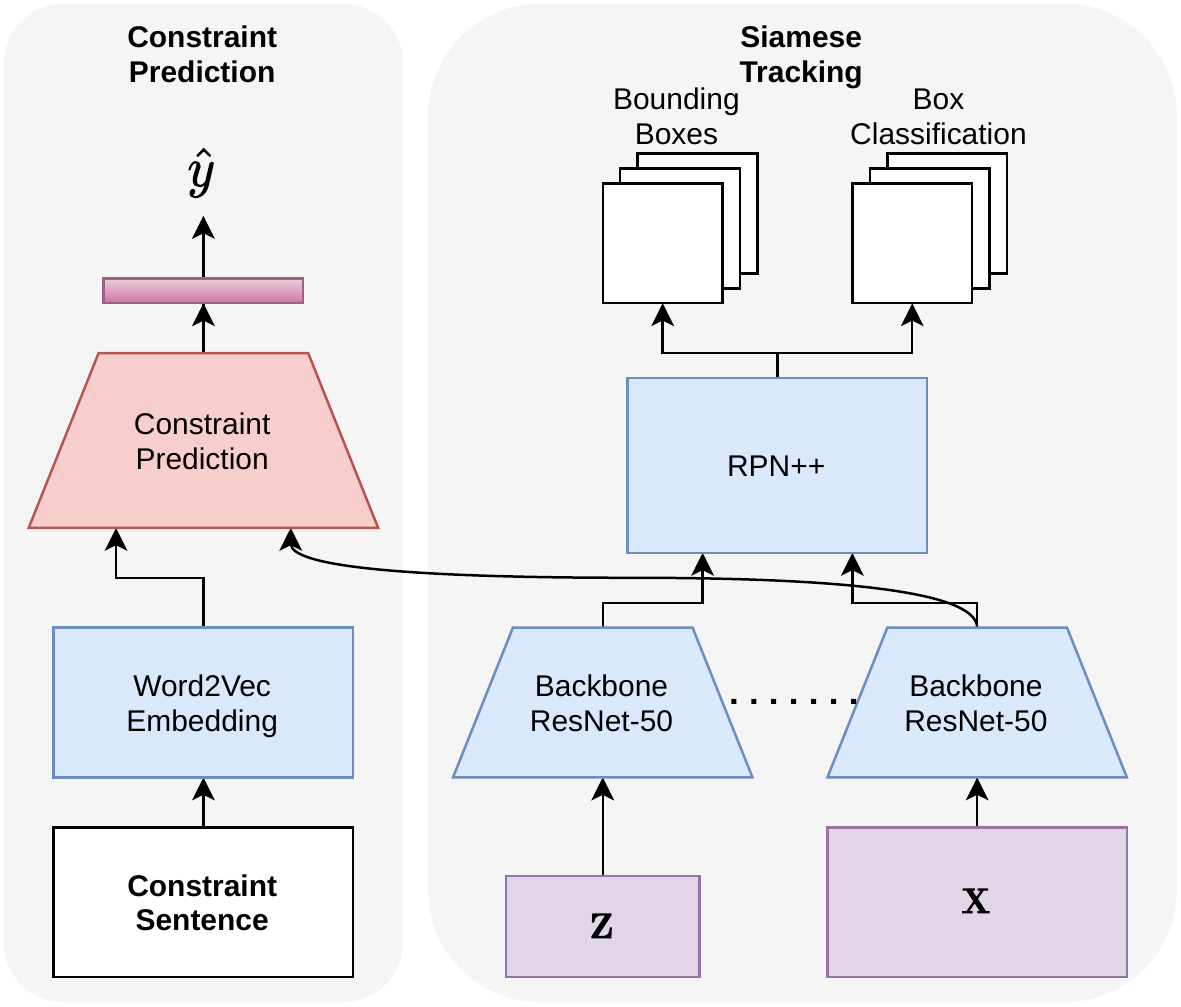}
     \caption{\textbf{Proposed Framework.}}
      \label{fig:model-setup}
  \end{subfigure}
  \hfill
    \begin{subfigure}[b]{0.34\textwidth}
        \centering
        \includegraphics[width=1\textwidth]{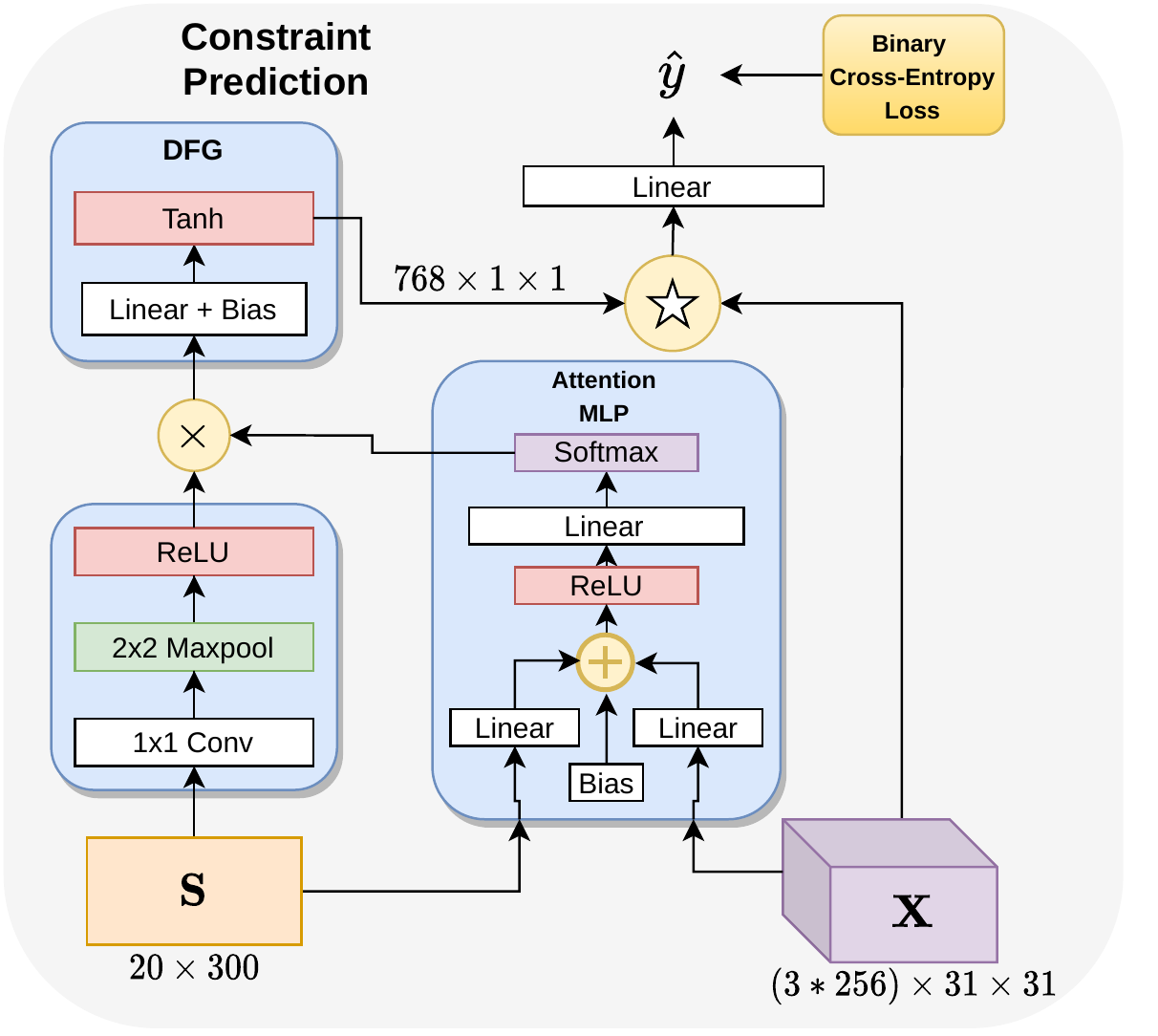}
        \caption{\textbf{\modelI{}.}}
        \label{fig:dynamic-conv-model}
    \end{subfigure}
    \hfill
    \begin{subfigure}[b]{0.27\textwidth}
      \centering
      \includegraphics[width=1\linewidth]{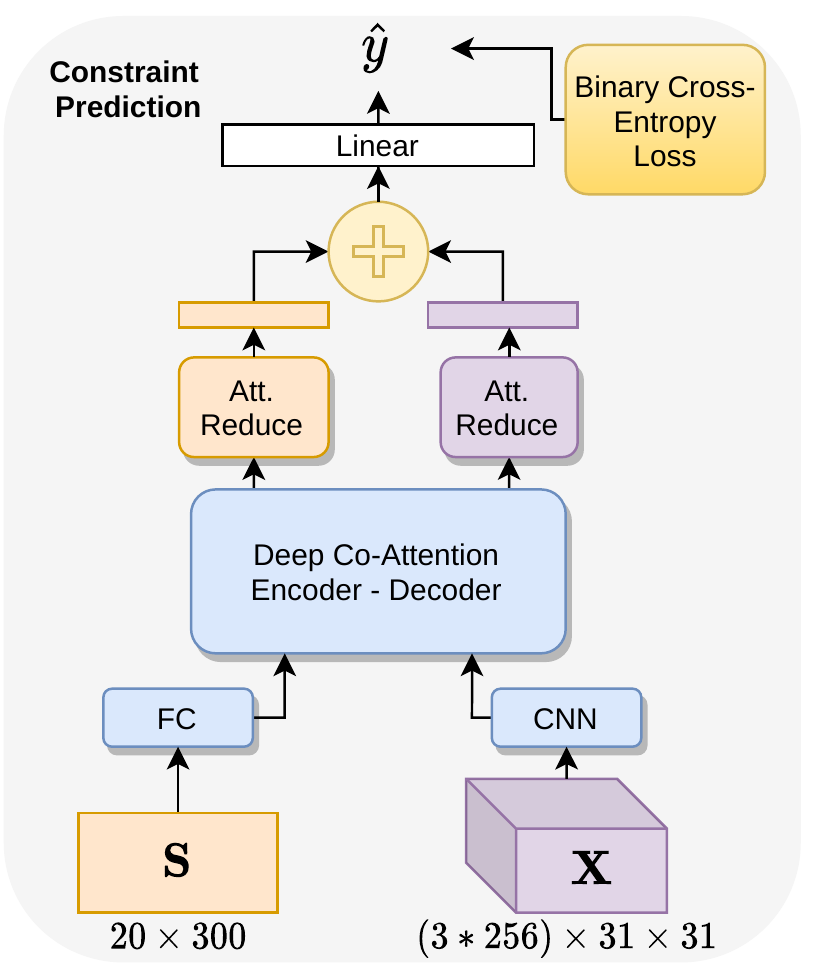}
      \caption{\textbf{\modelII{}.}}
      \label{fig:model_II}
    \end{subfigure}%
  \caption{
  \textbf{Proposed architectures.}  A high level overview of the proposed architecture is shown in figure \ref{fig:model-setup},
  the right hand side is the setup of a siamese tracker, and the left hand side shows the constraint prediction part of the model.
  The siamese tracker receives as input the search image $x^t$ of the target at time-step $t$ (frame $t$ of the video), using the ground truth reference image $z^{(t=0)}$ to match the target in time-step $t$ to the original image of the target. The constraint prediction part of the model receives as input a natural language sentence, and has to predict whether the frame $x^t$ matches the described constraint.
  Figures \ref{fig:dynamic-conv-model} and \ref{fig:model_II} showcase our proposed \modelI{} and \modelII{} respectively, which are an implementation of the red box in figure \ref{fig:model-setup}. 
  }
  \label{fig:three graphs}
\end{figure*}

\textbf{Word Embeddings.} \label{sec:word_embeddings}
To integrate the prediction of the constraint into the SiamRPN++ model, the natural language query is first processed using a word2vec model\footnote{Trained on the google news dataset.} (with $300$ dimensional embeddings), which is a common choice \cite{Li_2017_CVPR_Natural_Language, Gavrilyuk_2018_sentence}. 
After which the input sequence is padded to a fixed length $L = 20$.
The word embeddings will produce a feature map of $\bS \in \realnumbers^{20 \times 300}$, where we denote the input sentence as the word sequence $(w_1, ..., w_k)$.

\textbf{Constraint Prediction.} \label{sec:method_constraint_prediction}
Given the lingual constraint processed into a embedding $\bS \in \realnumbers^{20 \times 300}$ and the search image processed into a feature map by the backbone network $\varphi$, $\bX = \varphi(x) \in \realnumbers^{(3 \times 256) \times 31 \times 31}$ the constraint has to be predicted from these features.
Figure \ref{fig:model-setup} gives a high level overview\footnote{Note that this graph only shows a high-level overview, and hides the complexity in the individual modules for clarity.} of this proposed setup, in which the implementation for the constraint prediction box (coloured in red) is the part of the model our architectural research is concerned with.

The challenge for the constraint prediction is clearly the integration of the information in the lingual constraint features $\bS$  and that of the search image $\bX$.
In other words, the problem could also be formulated as `detection and classification of the lingual constraint in the search image'.
Since the models output a scalar $\hat{y_i}$ for the constraint prediction, we optimise the models using a binary cross-entropy loss, 
which is suitable for this binary machine learning problem.

\subsection{SiamCT-DFG}
Based on  the multi-modal tracking literature \cite{Gavrilyuk_2018_sentence, Li_2017_CVPR_Natural_Language, Hu2016SegmentationFN} and especially Li \etal \cite{Li_2017_CVPR_Natural_Language}, which is the closest work to our research, we construct a model for tracking with lingual constraints, displayed in figure \ref{fig:dynamic-conv-model}.


\textbf{Word Embeddings Processing.}
Instead of using a LSTM to process the word embeddings, Gavrilyuk \etal \cite{Gavrilyuk_2018_sentence} report improvements by keeping the processing of the word embeddings fully convolutional.
Following this, the word embeddings are fed through a 1D Convolution with padding, to keep the dimensions consistent, followed by a ReLU activation function and a 2D max-pool with a kernel and stride of size $2$.
This module transforms the word embeddings of $\bS \in \realnumbers^{20 \times 300}$ to a feature matrix $\bH = \text{CNN}(\bS) \in \realnumbers^{10 \times 150}$, and represents the resulting feature vector of two interpolated words as $\bh_i \in \realnumbers^{1 \times 150}$.

\textbf{Attention MLP.}
To integrate information from the search frame into the word embeddings, we adopt the attention MLP from Li \etal \cite{Li_2017_CVPR_Natural_Language}\footnote{note that this is a significantly different attention module compared to self-attention and co-attention}.
The attention MLP aims to create attention weights for the words in the lingual constraint that are the most important for the subsequent steps in the model.
It creates these attention weights based on the search frame and the word embeddings itself, the idea being to attend the constraint based on the words in the constraint that are also visible in the search frame.


\textbf{Dynamic Filters.}
Once the final representation of the word embeddings is acquired, dynamic convolutional filters \cite{Li_2017_CVPR_Natural_Language, Gavrilyuk_2018_sentence} are created from the sentence.
The dynamic convolution filters enable the model to create filters specific to the provided lingual constraint, thus producing activations specific to the (important) words present in the constraint.
Following the improvements made by Gavrilyuk \etal \cite{Gavrilyuk_2018_sentence}, the dynamic filter layer is a linear layer with a tanh activation function, that produces the convolutional filters,
$f = \tanh{ (\bW_f \widetilde{\bv} + \bb_f) } \in \realnumbers^{768 \times 1 \times 1}.$
Consequently, the filters $f$ are convolved with the features of the search frame $x$ to produce a activation map, $\bA = \bX \star f$, where $\star$ denotes a depth-wise cross correlation layer \cite{Li2018SiamRPNEO_siamRPN++}.
Finally, the activation map $\bA$ is flattened and projected to a scalar using a linear layer, after which the loss is calculated using the BCE loss as described earlier.

\subsection{\modelII{}} \label{subsec:model_II}
In this section, we introduce a second constraint prediction model that is based on state-of-the-art VQA literature and especially on the MCAN model \cite{VQA:Yu2019DeepMC}\footnote{For a thorough description of the MCAN model, we refer to the original paper.}.
A high-level overview of \modelII{} is presented in figure \ref{fig:model_II}.


\textbf{Preprocessing.}
Before the inputs are processed in the MCAN model, the feature dimensions of the word embeddings and the features of the search image $\bX$ have to be matched.
To achieve this, the word embeddings are fed through a Linear layer, (FC($d$)-ReLU), which expands the word embeddings to the feature dimension $d$ of the MCAN model, equal to the channel dimension of the image features $\bX$.
Yet, because of the computational complexity of the model, the number of features have to be further reduced given our limited resources.
This is achieved by reducing the number of feature locations of the image $\bX$, from $31 \times 31$ to $15 \times 15$, (Conv2d(d, d)-ReLU-MaxPool2d(k=2, s=3)). 
Consequently, the data is reshaped to the attention model shape (by concatenating the width and height dimensions of the image), in addition to another convolution, that is employed to further reduces the dimensionality of the image features from $15 \times 15 = 225$ to a dimensionality of $75$, (Conv1d(d, d, k=3, p=1, groups=8)-ReLU-MaxPool1d(k=3, s=3)).
Producing the reduced image features $\bX_r \in \realnumbers^{(256 \times 3) \times 75}$.

\textbf{Deep Co-Attention.}
Taking the reduced image features $\bX_r$ and the word embedding features of the constraint $\bS$, the MCAN model co-attends the image based upon the constraint, after it has self-attended and encoded the constraint.
After the deep co-attention model, the features have to be fused to one representation and classified.
This is achieved with the attentional reduction model \cite{VQA:Yu2019DeepMC}.
After the attentional reduction, the features are added together and projected to a single scalar using a linear layer, from which the BCE loss is calculated.


\textbf{Pyramid Pooling Module.}
To further improve the localisation of the constraint, we also incorporate the use of the Pyramid Pooling Module (PPM) which was originally proposed by  Zhao \etal \cite{Zhao2017PyramidSP}, and now successfully integrated into VQA with significant performance boosts shown by Jiang \etal \cite{VQA:jiang2020defense}.
The PPM module adds global features at every location in the image, computed at different scales $s$.
The function of these additional channels can be interpreted as a kind of prior for the image.

\section{Experiments} \label{sec:experiments}  
To evaluate the proposed models for tracking with lingual constraints, we conduct quantitative as well as qualitative experiments on the newly introduced datasets, c-LaSOT and c-MOT16.
We use a pre-trained SiamRPN++ tracking backbone in our models and only train the constraint prediction modules.
For our experiments, we conduct an analysis on \modelI{}, and \modelII{}, which we compare qualitatively and quantitatively.
For our ablation study, we add two extra models: \modelIIppm{} features the additional Pyramid Pooling Module, conversely \modelIatt{} does not feature the Attention MLP, and we study the impact of these changes accordingly.
Since there is no research that features a baseline that we can compare to, we utilise a majority baseline in our experiments. 

\textbf{Training Details.}
For training, the models are first pre-trained on the COCO dataset for $20$ epochs (see section \red{D}), and subsequently trained and evaluated on c-MOT16 or c-LaSOT.
For the optimiser, we employ SGD and ADAM \cite{OPTIM:Kingma2015AdamAM} for \modelI{} and \modelII{} respectively \cite{OT:Popel2018TrainingTF, VQA:Yu2019DeepMC}.
Further training specifics, hyperparameters, and hardware details are further specified in the appendix, in section \red{E}.

\textbf{Constraint Prediction Evaluation.} \label{sec:evaluation}
The model outputs a single number, $p \in [0, 1]$, which we can interpret as the probability of the constraint being positive. 
Afterwards, this probability will have to be 'rounded' to $\hat{y} \in \{0, 1\}$, called thresholding, influencing the trade-off between precision and recall. 
Since the goal is to find the clips in a video where the constraint is satisfied, it is crucial that the model does not predict too many false positives.
On the other hand, if whole sequences in the video do not receive a positive prediction (which are false-negative predictions), these clips will be missed. 
In contrast, if at least one frame in the clip is classified positively, the user of the model should then be able to find the rest of the frames in this clip, where the constraint is satisfied.
We conclude that precision is more important than recall, and thus evaluate the experimental results with a $F$-measure of $F_{0.5}$.
In addition, given the pitfalls of the $F$-score, we report the MCC, AUC and AP score as well.

\subsection{Results}
\begin{figure}
  \centering
  \begin{subfigure}[b]{0.48\textwidth}
     \centering
     \includegraphics[width=1\textwidth]{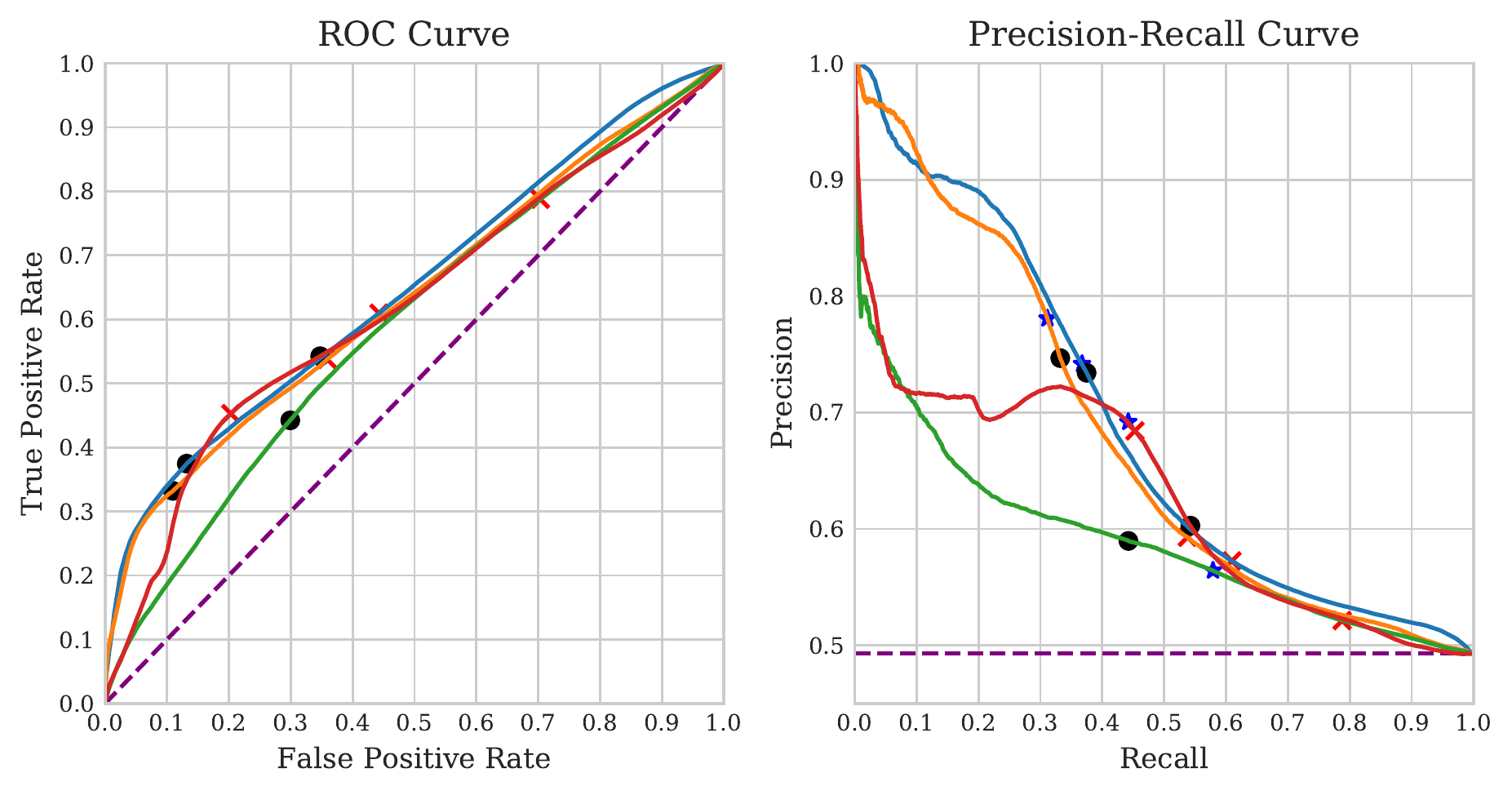}
     \caption{\textbf{c-MOT16.}}
      \label{fig:mot_threshold_curves}
  \end{subfigure}
  \begin{subfigure}[b]{0.48\textwidth}
        \centering
        \includegraphics[width=1\textwidth]{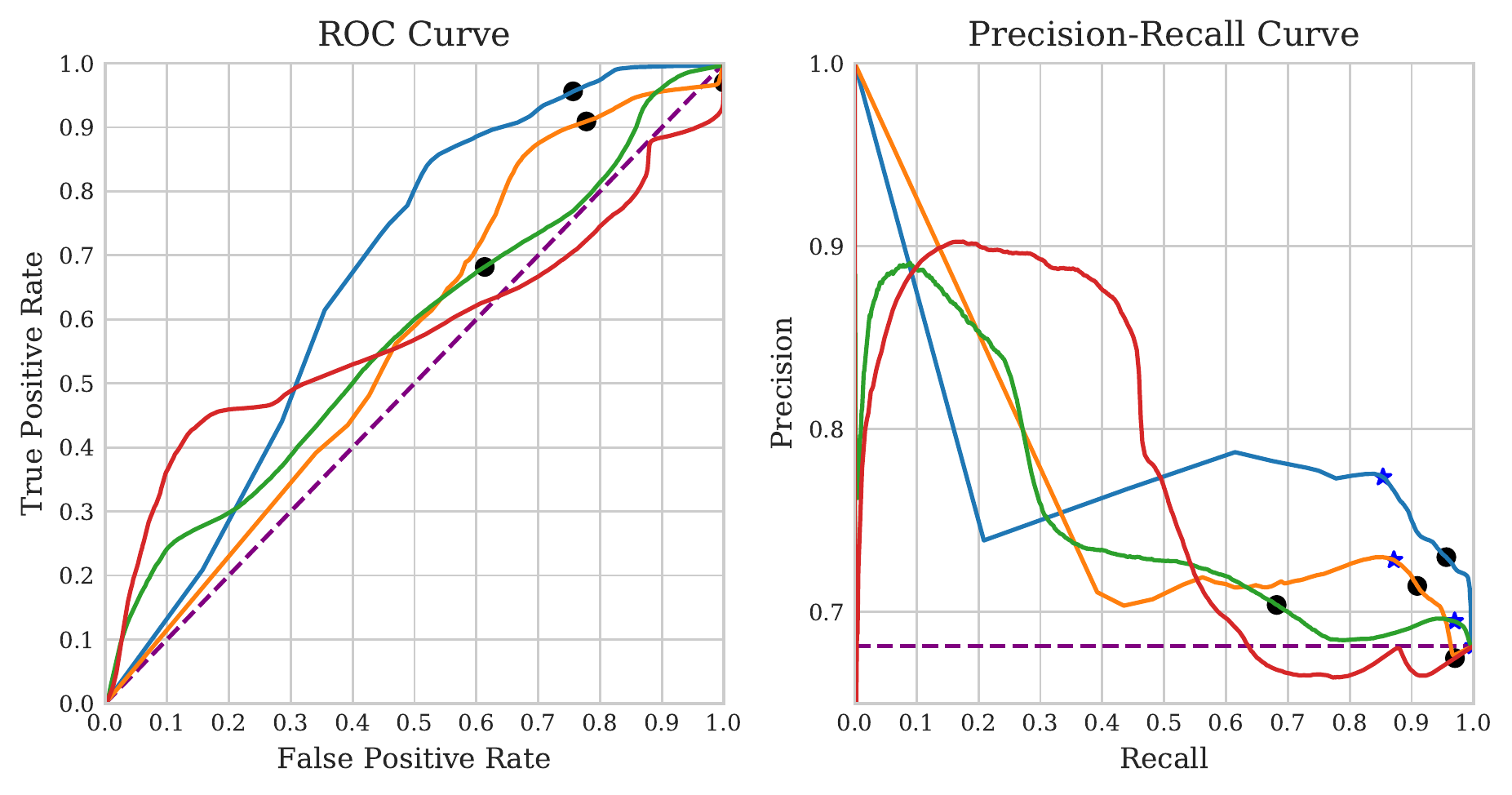}
        \caption{\textbf{c-LaSOT.}}
        \label{fig:lasot_threshold_curves}
  \end{subfigure}
  \caption{\textbf{Performance plots}, for c-MOT16 and c-LaSOT datasets showing the thresholding trade-off for each model. 
  The legend goes as follows: Positive baseline (\purple{\dashed}), \modelIIppm{} (\blue{\full}), \modelII{}  (\orange{\full}),  \modelI{} (\green{\full}), \modelIatt{} (\darkred{\full}).
  The star (\blue{$\bigstar$}) marks the threshold where the $F_{0.5}$ score is the highest for the model, and the \red{\Xmark} marks the optimal threshold calculated on the validation dataset. Lastly, the \tikzcircle{2pt} marks the score for a default $\frac{1}{2}$ threshold.
  }
  \label{fig:res_threshold_curves}
\end{figure}

Visually the main results on the c-MOT16 and c-LaSOT dataset are shown in figure \ref{fig:res_threshold_curves}, displaying ROC curves and Precision-Recall curves. The same results are reported numerically in table \ref{tab:mot-lasot-quant-results}.

For the c-MOT16 dataset, the most outstanding result in figure \ref{fig:mot_threshold_curves} is that \modelI{} has the lowest performance of the four models.
In contrast to these results, we see that the other models are all very close in performance.
Secondly, \modelIIppm{} outperforms all other models slightly everywhere, except for a small interval for both curves.

In terms of thresholds in the Precision-Recall curve, we observe that the optimal $F_{0.5}$ threshold is close to the default threshold of $0.5$.
Furthermore, the thresholds computed on the validation set are far away from the optimal threshold for all but one model: \modelIatt{} has a validation threshold almost equal to its optimal threshold.
This will significantly improve its performance on metrics that do not take the threshold into account, especially in comparison with the other models.
This aforementioned fact makes a comparison between the models based on the threshold insensitive metrics difficult and potentially misleading.
In terms of the threshold insensitive metrics (AP, ROC AUC), \modelIIppm{} clearly performs the best.
For average precision (AP) the model scores $68.6$, $1.3$ higher then the second best, and almost $5$ points higher than the third best.

\begin{figure*}[hbtp!]
    \centering
\begin{subfigure}[t]{\dimexpr0.22\textwidth+20pt\relax}
    \includegraphics[width=\dimexpr\linewidth-20pt\relax]
    {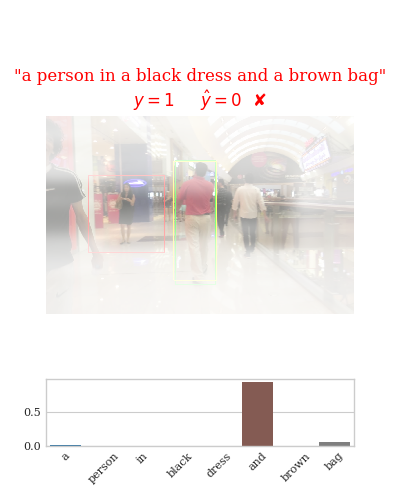}
    
     \includegraphics[width=\dimexpr\linewidth-20pt\relax, height=3.5cm]
     {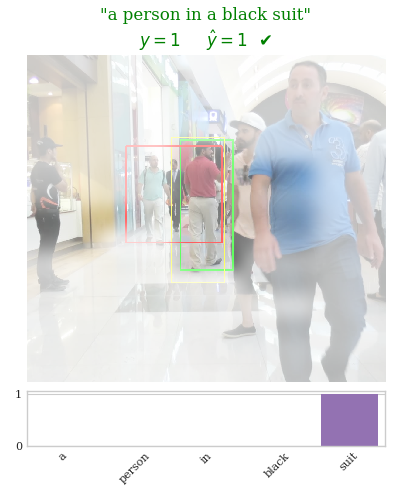}

     \includegraphics[width=\dimexpr\linewidth-20pt\relax]{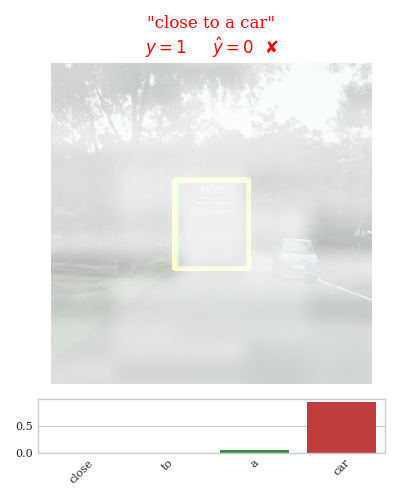}
     
     \includegraphics[width=\dimexpr\linewidth-20pt\relax]{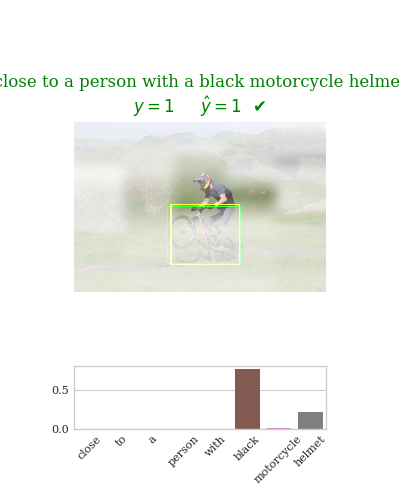}
    
     \caption{\textbf{\modelIIppm{}}}
\end{subfigure}\hfill
\begin{subfigure}[t]{0.22\textwidth}
    \includegraphics[width=\textwidth]  
    {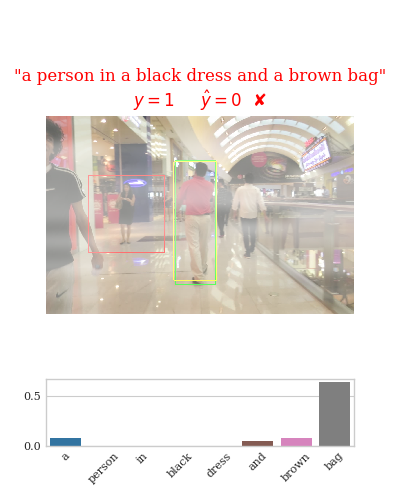}
    
    \includegraphics[width=\textwidth, height=3.5cm]
    {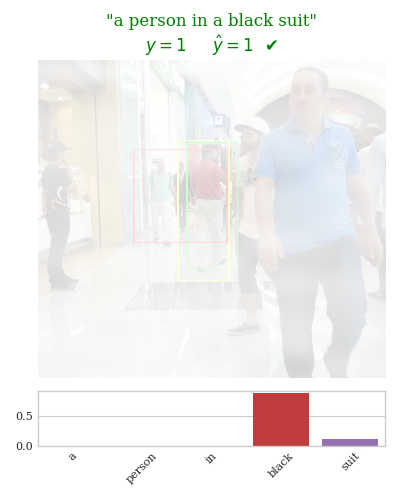}

    \includegraphics[width=\textwidth]
    {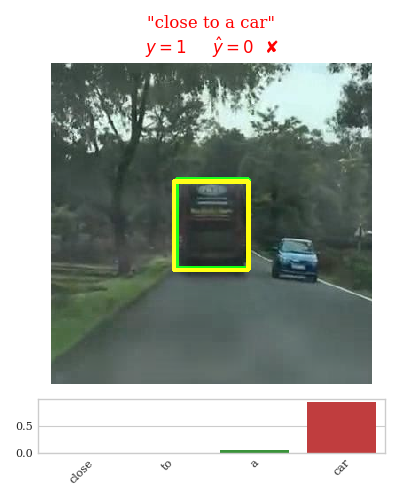}
    
    \includegraphics[width=\textwidth]
    {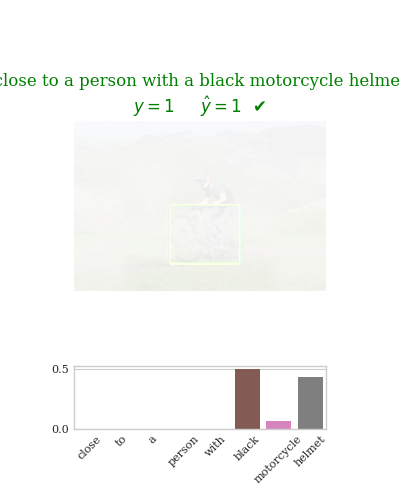}
    
    \caption{\textbf{\modelII{}}}
\end{subfigure}\hfill
\begin{subfigure}[t]{0.22\textwidth}
    \includegraphics[width=\textwidth]  
    {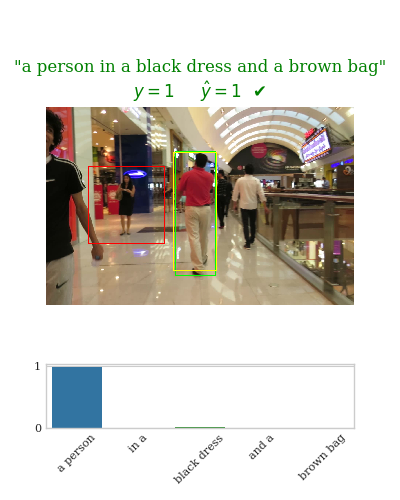}
    
    \includegraphics[width=\textwidth, height=3.5cm]
    {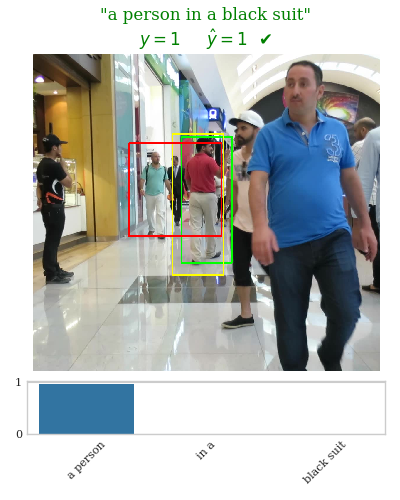}

    \includegraphics[width=\textwidth]
    {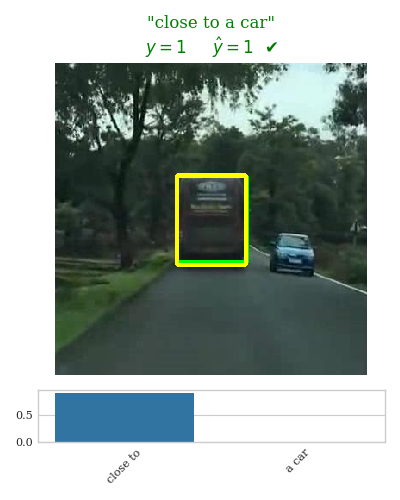}
    
    \includegraphics[width=\textwidth]
    {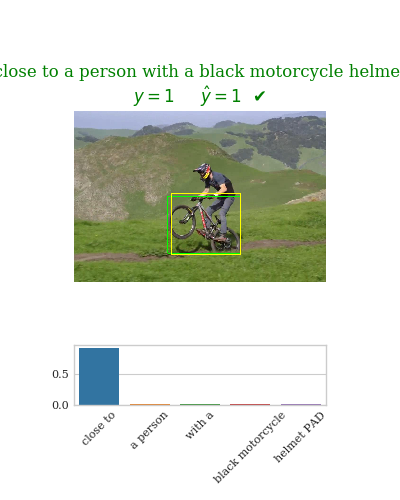}
    
    \caption{\textbf{\modelI{}}}
\end{subfigure}\hfill
\begin{subfigure}[t]{0.22\textwidth}
    \includegraphics[width=\textwidth]  
    {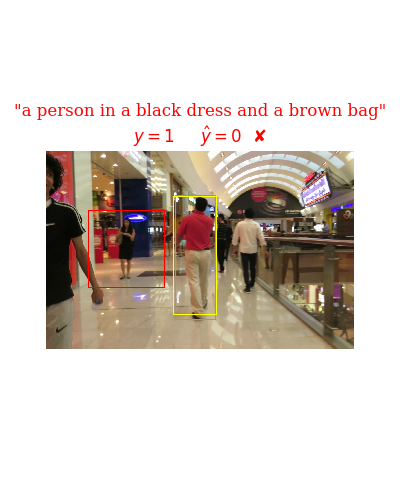}
    
    \includegraphics[width=\textwidth, height=3.5cm]
    {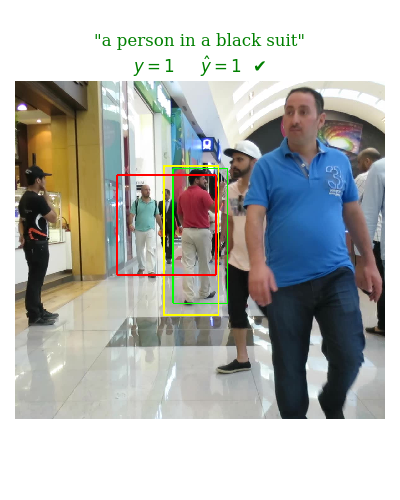}

    \includegraphics[width=\textwidth]
    {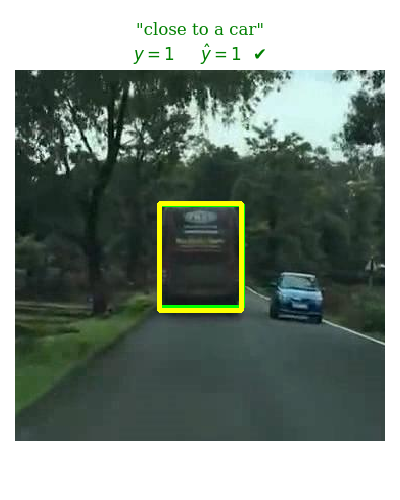}
    
    \includegraphics[width=\textwidth]
    {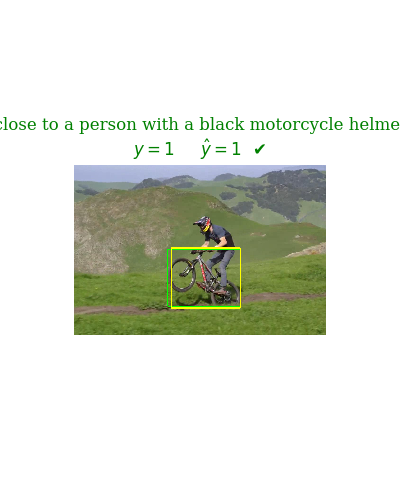}
    
    \caption{\textbf{\modelIatt{}}}
\end{subfigure}
    \caption{\textbf{Visual predictions on the c-MOT16 (first 2 rows) and c-LaSOT (last 2 rows) datasets.} The rows depict a typical frame from the dataset, and the columns show the predictions of a model on this particular frame. For the deep co-attention models (\modelII{}), the final image attention is visualised, where the image is more clear in the attended parts. The lingual attention weights are visualized (when applicable), and depicted by a bar chart below the frame. The attention for \modelI{} is on word pairs, as the word embeddings are processed with a $1\times1$ convolution before being attended. The sentence above the frames denote the lingual constraint, with below it the predictions, where $y$ and $\hat{y}$ denote the groundtruth and prediction respectively.}
    \label{fig:mot_visual_predictions}
\end{figure*}

Next, we present visual depictions of the predictions for typical examples of the c-MOT16 dataset and the LaSOT datasets in figure \ref{fig:mot_visual_predictions}, and show the attention (where applicable) on the words in the constraint and on the visual frame.
Looking at the attention distributions on the lingual constraints, it stands out that the attention for the different models does not capture all the essential words in the sentences.
For instance, \modelI{} attends the words with the arguably the least information, which do not hold vital information for recognising the constraint objects in the frame.
In terms of lingual attention, \modelII{} has the best overall attention, showing more well rounded attention on words of importance.
If we now turn to the co-attention of the frames (\modelII{}, \modelIIppm{}) based upon the lingual constraint, it is clear that the visual attention is not able to directly attend the objects in the sentence.
We expect the visual attention to be focused on the words that were strongly attended in the sentence, however, there are no examples visible where the model specifically attended the objects in the frame that were also attended in the sentence. We present further results and attention maps in the supplementary materials, in section \ref{sec:appendix_mot_results} and \ref{sec:appendix_lasot_results}.

\begin{table*}
\centering
\caption{\textbf{Quantitative results}. Results are shown for both the c-MOT16 and c-LaSOT datasets. The most important metrics are shown first, AP and ROC AUC, after which the metrics that are dependant on the calibration threshold are shown. First the results based upon the threshold computed on the validation set are reported, denoted by $\dagger$ (the c-LaSOT dataset has no validation set), after which the results using the optimal threshold determined on the test set itself, marked by a $*$.}
\label{tab:mot-lasot-quant-results}
\resizebox{\textwidth}{!}{%
\begin{tabular}{@{}lllllllllll@{}}
\toprule
Dataset & Model & AP & ROC AUC & $F_{0.5}^{\dagger}$ & $MCC^{\dagger}$ & ${\dagger}$ Val Threshold & $*$ Optimal Threshold & Threshold Diff & $F_{0.5}^{*}$ & $MCC^{*}$ \\ \midrule
\multirow{4}{*}{c-MOT16} & \modelI{}  & 59.0          & 59.4          & 55.9          & 9.7           & 0.393 & 0.460 & 0.067          & 56.7          & 14.4          \\
                         & \modelIatt{} & 62.7          & 62.2          & \textbf{62.1} & \textbf{26.6} & 0.569 & 0.580 & \textbf{0.011} & \textbf{62.2} & 27.0          \\
                         & \modelII{} & 67.5          & 63.4          & 58.1          & 18.0          & 0.277 & 0.542 & 0.265          & 60.0          & \textbf{28.5} \\
                         & \modelIIppm{}  & \textbf{68.6} & \textbf{64.9} & 58.0          & 16.7          & 0.273 & 0.513 & 0.240          & 61.6          & 28.3          \\ \midrule
\multirow{4}{*}{c-LaSOT} & \modelI{} & 75.3          & 57.3          & 69.9          & 6.8           & 0.5   & 0.367 & -0.133         & 69.9          & 6.8           \\
                         & \modelIatt{} & \textbf{77.6} & 57.9          & 71.8          & -9.9          & 0.5   & 0.430 & \textbf{-0.07} & 71.8          & -9.9          \\
                         & \modelII{} & 71.2          & 56.9          & 74.6          & 18.0          & 0.5   & 0.717 & 0.217          & 74.6          & 18.0          \\
                         & \modelIIppm{} & 76.4          & \textbf{66.7} & \textbf{76.6} & \textbf{30.1} & 0.5   & 0.994 & 0.494          & \textbf{76.6} & \textbf{30.1} \\ \bottomrule
\end{tabular}%
}
\end{table*}

\subsection{Ablation Study}
Next, we do a in-depth comparison of our models, and identify the key takeaways.  

\textbf{\modelII{}.}
When comparing \modelIIppm{} to \modelII{}, the PPM improved the AP and the ROC AUC score for both data sets considerably.
Furthermore, in the threshold plots, it performs as well or better for every possible threshold, showing that the module does not make a trade-off in terms of precision and recall.
When comparing the two models qualitatively, the \modelIIppm{} module seems to have slightly better guided attention on the images.
In addition, in the guided attention layers, \modelIIppm{} has the attention more distributed over several heads, and is also less concentrated per head.
Since the PPM module adds spatial features to the image map that go past the boundaries of that location of the feature map, it seems that the \modelI{} able to divide its attention more easily to important parts of the feature maps.
However, since the differences are minimal, this cannot be declared significant.
In contrast, when comparing the two models in terms of attention on the lingual constraint, \modelII{} seems to divide its attention over slightly more words\footnote{It is however unknown whether this is an important attribute to arrive at accurate predictions.}. 

\textbf{\modelI{}.}
One difficulty of evaluating \modelIatt{}, is that it is not interpretable to the same level as the models that have attention, thus we can mainly analyse and compare its predictive power to that of \modelI{}.
Surprisingly, \modelIatt{} scores higher on almost all metrics and similarly has `better' Precision-Recall curves and ROC AUC curves.
However, out of the two models, it is clear that \modelIatt{} inhibits more variance in its results, depending on the threshold, which could be an indication of overfitting.
Having said that, it is evident that \modelI{} does not benefit from the `Attention MLP' that it features, particularly from the fact that the attention does not attend the right words.

\textbf{Overall comparison.}
In terms of quantitative results, it is clear that \modelIIppm{} performs the most consistent of all the models, scoring the highest on the AP and ROC AUC metrics in 3 out of 4 cases.
However, \modelIatt{} does show good performance in some parts of the threshold curves, but, it does not score as well on the MOT dataset, especially at a lower recall and higher precision value.
A reason for this could be the fact that \modelIatt{} misses the sophisticated attention mechanism present in \modelII{}. 
This mechanism could be required for the more complex lingual constraints present in the c-MOT16 dataset, compared to the c-LaSOT dataset.
However, a possible advantage of \modelIatt{} over \modelIIppm{} is that it features a (classically) more powerful vision-based architecture, that could enable it to more easily identify the target, as seen from the good performance on the c-LaSOT dataset.

\section{Conclusion}
In this research, we have introduced a new research topic, which brings visual object tracking closer to (many) real-world applications.
Lingually constrained tracking is an exciting new research direction, as it offers the opportunity to bring together many other flourishing fields in artificial intelligence, like tracking, natural language processing, video analysis, object recognition, and many more.

This study set out to establish how a deep tracker could efficiently be extended to incorporate lingual constraints.
In conclusion, the findings clearly indicate that extending the SiamRPN++ model with the MCAN model is a promising direction.
The research has also shown that the Pyramid Pooling Module significantly improves the results of the MCAN model for predicting the constraint.
Moreover, this study has examined the role of the `Attention MLP' \cite{Li_2017_CVPR_Natural_Language} for lingually constrained tracking, and the findings clearly indicate that this module is not an important component for future work.

A limitation of this study is the variability of the data that was learned and experimented with, due to the somewhat limited amount of annotated data for said task.
In an effort to improve this, we have introduced the c-LaSOT, and c-MOT16 datasets, which has an abundance of tracks (tracking sequences); however, it is constructed of only nine videos.
This data restriction consequently increased the variability of the overall results.
Future studies should focus on introducing new datasets for lingually constrained tracking, such that more progress can be made in this research direction.
Furthermore, more complex lingual constraints could be incorporated given more data, in turn enabling new directions of research, like including actions into the lingual constraints. 

\clearpage


\begin{thebibliography}{10}\itemsep=-1pt

\bibitem{Bertinetto2016FullyConvolutionalSN_siamFCv1}
Luca Bertinetto, Jack Valmadre, Joao~F Henriques, Andrea Vedaldi, and Philip~HS
  Torr.
\newblock Fully-convolutional siamese networks for object tracking.
\newblock In {\em European conference on computer vision}, pages 850--865.
  Springer, 2016.

\bibitem{Bhat2019LearningDM_DIMP}
Goutam Bhat, Martin Danelljan, Luc~Van Gool, and Radu Timofte.
\newblock Learning discriminative model prediction for tracking.
\newblock In {\em Proceedings of the IEEE/CVF International Conference on
  Computer Vision (ICCV)}, 10 2019.

\bibitem{Danelljan2016ECOEC}
Martin Danelljan, Goutam Bhat, Fahad~Shahbaz Khan, and Michael Felsberg.
\newblock Eco: Efficient convolution operators for tracking.
\newblock {\em 2017 IEEE Conference on Computer Vision and Pattern Recognition
  (CVPR)}, pages 6931--6939, 2016.

\bibitem{Danelljan2018ATOMAT}
Martin Danelljan, Goutam Bhat, Fahad~Shahbaz Khan, and Michael Felsberg.
\newblock Atom: Accurate tracking by overlap maximization.
\newblock In {\em Proceedings of the IEEE/CVF Conference on Computer Vision and
  Pattern Recognition (CVPR)}, 6 2019.

\bibitem{OT:Dietterich1996ApproximateST}
Thomas~G Dietterich.
\newblock {Approximate Statistical Tests for Comparing Supervised
  Classification Learning Algorithms}.
\newblock {\em Neural Computation}, 10(7):1895--1923, 1998.

\bibitem{DS:Fan2018LaSOTAH}
Heng Fan, Liting Lin, Fan Yang, Peng Chu, Ge Deng, Sijia Yu, Hexin Bai, Yong
  Xu, Chunyuan Liao, and Haibin Ling.
\newblock Lasot: A high-quality benchmark for large-scale single object
  tracking.
\newblock In {\em Proceedings of the IEEE/CVF Conference on Computer Vision and
  Pattern Recognition (CVPR)}, 6 2019.

\bibitem{VQA:Farazi2020AccuracyVC}
Moshiur~R. Farazi, Salman~Hameed Khan, and Nick Barnes.
\newblock Accuracy vs. complexity: A trade-off in visual question answering
  models.
\newblock {\em ArXiv}, abs/2001.07059, 2020.

\bibitem{Gavrilyuk_2018_sentence}
Kirill Gavrilyuk, Amir Ghodrati, Zhenyang Li, and Cees G.~M. Snoek.
\newblock Actor and action video segmentation from a sentence.
\newblock In {\em Proceedings of the IEEE Conference on Computer Vision and
  Pattern Recognition (CVPR)}, 6 2018.

\bibitem{VQA:Gokhale2020VQALOLVQ}
Tejas Gokhale, Pratyay Banerjee, Chitta Baral, and Yezhou Yang.
\newblock Vqa-lol: Visual question answering under the lens of logic.
\newblock {\em ArXiv}, abs/2002.08325, 2020.

\bibitem{Guo2017LearningDSiam}
Qing Guo, Wei Feng, Ce Zhou, Rui Huang, Liang Wan, and Song Wang.
\newblock Learning dynamic siamese network for visual object tracking.
\newblock {\em 2017 IEEE International Conference on Computer Vision (ICCV)},
  pages 1781--1789, 2017.

\bibitem{Hu2016SegmentationFN}
Ronghang Hu, Marcus Rohrbach, and Trevor Darrell.
\newblock Segmentation from natural language expressions.
\newblock In {\em European Conference on Computer Vision}, pages 108--124.
  Springer, 2016.

\bibitem{VQA:jiang2020defense}
Huaizu Jiang, Ishan Misra, Marcus Rohrbach, Erik Learned-Miller, and Xinlei
  Chen.
\newblock In defense of grid features for visual question answering.
\newblock {\em IEEE Conference on Computer Vision and Pattern Recognition
  (CVPR)}, 2020.

\bibitem{OPTIM:Kingma2015AdamAM}
Diederik Kingma and Jimmy Ba.
\newblock Adam: A method for stochastic optimization.
\newblock {\em International Conference on Learning Representations}, 12 2014.

\bibitem{VOT2017:kristan2017visual}
Matej Kristan, Ales Leonardis, Jiri Matas, Michael Felsberg, Roman Pflugfelder,
  Luka Cehovin~Zajc, Tomas Vojir, Gustav Hager, Alan Lukezic, Abdelrahman
  Eldesokey, et~al.
\newblock The visual object tracking vot2017 challenge results.
\newblock In {\em Proceedings of the IEEE international conference on computer
  vision workshops}, pages 1949--1972, 2017.

\bibitem{Li2018SiamRPNEO_siamRPN++}
Bo Li, Wei Wu, Qiang Wang, Fangyi Zhang, Junliang Xing, and Junjie Yan.
\newblock Siamrpn++: Evolution of siamese visual tracking with very deep
  networks.
\newblock In {\em Proceedings of the IEEE/CVF Conference on Computer Vision and
  Pattern Recognition (CVPR)}, 6 2019.

\bibitem{Li2018HighPV_siamRPN}
Bo Li, Junjie Yan, Wei Wu, Zheng Zhu, and Xiaolin Hu.
\newblock High performance visual tracking with siamese region proposal
  network.
\newblock {\em 2018 IEEE/CVF Conference on Computer Vision and Pattern
  Recognition}, pages 8971--8980, 2018.

\bibitem{Li_2017_CVPR_Natural_Language}
Zhenyang Li, Ran Tao, Efstratios Gavves, Cees G.~M. Snoek, and Arnold~W.M.
  Smeulders.
\newblock Tracking by natural language specification.
\newblock In {\em The IEEE Conference on Computer Vision and Pattern
  Recognition (CVPR)}, 7 2017.

\bibitem{DS:Lin2014MicrosoftCOCO}
Tsung-Yi Lin, Michael Maire, Serge Belongie, James Hays, Pietro Perona, Deva
  Ramanan, Piotr Doll{\'a}r, and C~Lawrence Zitnick.
\newblock Microsoft coco: Common objects in context.
\newblock In {\em European conference on computer vision}, pages 740--755.
  Springer, 2014.

\bibitem{VQA:Lu2016HierarchicalQC}
Jiasen Lu, Jianwei Yang, Dhruv Batra, and Devi Parikh.
\newblock Hierarchical question-image co-attention for visual question
  answering.
\newblock In D. Lee, M. Sugiyama, U. Luxburg, I. Guyon, and R. Garnett,
  editors, {\em Advances in Neural Information Processing Systems}, volume~29,
  pages 289--297. Curran Associates, Inc., 2016.

\bibitem{VQA:Martins2020SparseAS}
Pedro~Henrique Martins, Vlad Niculae, Zita Marinho, and Andr{\'e} F.~T.
  Martins.
\newblock Sparse and structured visual attention.
\newblock {\em ArXiv}, abs/2002.05556, 2020.

\bibitem{DS:milan2016mot16}
Anton Milan, Laura Leal-Taix{\'e}, Ian Reid, Stefan Roth, and Konrad Schindler.
\newblock Mot16: A benchmark for multi-object tracking.
\newblock {\em arXiv preprint arXiv:1603.00831}, 2016.

\bibitem{OT:numpy_oliphant2006guide}
Travis~E Oliphant.
\newblock {\em A guide to NumPy}, volume~1.
\newblock Trelgol Publishing USA, 2006.

\bibitem{OT:paszke2017automatic_pytorch}
Adam Paszke, Sam Gross, Soumith Chintala, Gregory Chanan, Edward Yang, Zachary
  DeVito, Zeming Lin, Alban Desmaison, Luca Antiga, and Adam Lerer.
\newblock Automatic differentiation in pytorch.
\newblock In {\em NIPS-W}, 2017.

\bibitem{OT:Popel2018TrainingTF}
Martin Popel and Ondrej Bojar.
\newblock Training tips for the transformer model.
\newblock {\em The Prague Bulletin of Mathematical Linguistics}, 110:43 -- 70,
  2018.

\bibitem{Valmadre_2017_CVPR_siamFCv2}
Jack Valmadre, Luca Bertinetto, Joao Henriques, Andrea Vedaldi, and Philip
  H.~S. Torr.
\newblock End-to-end representation learning for correlation filter based
  tracking.
\newblock In {\em The IEEE Conference on Computer Vision and Pattern
  Recognition (CVPR)}, 7 2017.

\bibitem{OT:numpy_van2011numpy}
Stefan Van Der~Walt, S~Chris Colbert, and Gael Varoquaux.
\newblock The numpy array: a structure for efficient numerical computation.
\newblock {\em Computing in Science \& Engineering}, 13(2):22, 2011.

\bibitem{Vaswani2017AttentionIA}
Ashish Vaswani, Noam Shazeer, Niki Parmar, Jakob Uszkoreit, Llion Jones,
  Aidan~N Gomez, {\L}ukasz Kaiser, and Illia Polosukhin.
\newblock Attention is all you need.
\newblock In {\em Advances in neural information processing systems}, pages
  5998--6008, 2017.

\bibitem{VQA:Yu2019MultimodalUA}
Zhou Yu, Yuhao Cui, Jun Yu, Dacheng Tao, and Qi Tian.
\newblock Multimodal unified attention networks for vision-and-language
  interactions.
\newblock {\em ArXiv}, abs/1908.04107, 2019.

\bibitem{VQA:Yu2019DeepMC}
Zhou Yu, Jun Yu, Yuhao Cui, Dacheng Tao, and Qi Tian.
\newblock Deep modular co-attention networks for visual question answering.
\newblock {\em 2019 IEEE/CVF Conference on Computer Vision and Pattern
  Recognition (CVPR)}, pages 6274--6283, 2019.

\bibitem{Zhao2017PyramidSP}
Hengshuang Zhao, Jianping Shi, Xiaojuan Qi, Xiaogang Wang, and Jiaya Jia.
\newblock Pyramid scene parsing network.
\newblock {\em 2017 IEEE Conference on Computer Vision and Pattern Recognition
  (CVPR)}, pages 6230--6239, 2017.

\end{thebibliography}

\clearpage
\appendix
\section*{Supplementary Material}
\section{Comparison to Li \etal \cite{Li_2017_CVPR_Natural_Language}} \label{sec:appendix_li_comparison}
Within the field, the work of Li \etal \cite{Li_2017_CVPR_Natural_Language}, is the most relevant to our research.
However, there are a few key differences when comparing their objective to our proposed task.
Li \etal \cite{Li_2017_CVPR_Natural_Language}, propose to track by natural language specification, and thus could track a target with a sentence of `a man with a dark backpack'.
Yet, if the target takes off their backpack at some point in the video, the attention part of their model will ensure that the words are ignored if they do no longer describe the target.
In reverse, if the target does not conform to the full description (in the given example the target would not have a backpack on), either the tracking would fail or the tracker would ignore the description about the object.
This is the opposite of what we are trying to achieve in tracking with visual object constraints, where the model may not ignore parts of the sentence, and instead predict when the sentence is fully satisfied in the video.



In the problem described in our paper, the tracker tracks by a starting specification of the target and has to predict when the constraint is satisfied.
The work of Li \etal \cite{Li_2017_CVPR_Natural_Language}, could be employed is as an one-shot detection model, which will only track the target when the constraint is satisfied.
However, a limitation of stopping tracking when the constraint is unsatisfied is that the target is lost in these frames, in other words, the tracker has to search through the whole frame whenever the constraint is not satisfied. 
Another limitation lies in the case of similar looking target objects (e.g., the tracking of a specific person when he has the ball in a football match).
Unless the sentence specifies the target very clearly, the model has no means to find the right target.
This is an inherent limitation of specifying the target via a natural language specification.
Since the work of Li \etal, was the first paper to integrate tracking and natural language queries, their work is taken as a take-off point for this work.

\section{Constrained LaSOT Dataset Details} \label{sec:appendix_ds_lasot}
In this section, we specify additional details of the c-LaSOT dataset, including how the dataset was annotated, and further specifications on the details of the constraints.
Annotating the videos was carried out by manually surveying the videos in the dataset and monitoring for any videos that have frames that contain any of the constraining object classes.
Once these videos were found, the beginning and endpoints of these time intervals were marked, where a constraining object was close to the target object. 
This results in one \textit{constraint track} per object constraint for each video, with intervals where the constraint was satisfied.
As a result one video can have multiple \textit{constraint tracks} if the tracking target happens to come in close proximity to different constraining objects, where `in close proximity' is defined as the object being within bounds for 50\% of $1.5$ times the width and the height of the target object itself.
Since there are no bounding-boxes for these constraining objects in the LaSOT dataset, these proximity checks were performed with an `eye test'.
Lastly, since the LaSOT dataset was meant as a first dataset, the constraint is defined as a sentence describing the constraining object, where only the name of the object (of the words in the lingual constraint) is important to determine whether the constraint is satisfied in a particular frame. 
\begin{table}
\centering
\caption{The attributes of one row in the annotations of the extended LaSOT dataset. Note that one \textit{constraint track} can have multiple rows associated with it, such that, in different time intervals in the same video, the constraint is satisfied.}
\label{tab:annotations}
\resizebox{0.45\textwidth}{!}{%
\begin{tabular}{ll}
\hline
Attribute         & Description                                                         \\ \hline
constraint\_track & The identifier of the constraint track.                             \\
sequence\_id      & The identifier of the LaSOT sequence of the video.                  \\
category          & The name of the LaSOT category (e.g., `bear' or `airplane')         \\
constraint\_from  & The frame number (inclusive) from when the constraint is satisfied. \\
constraint\_till  & The frame number (inclusive) till when the constraint is satisfied. \\
sentence          & The lingual constraint.                                            \\ \hline
\end{tabular}%
}
\end{table}

\subsection{Challenges} \label{sec:ds_lasot_challenges}
One problem with annotating the LaSOT dataset with language constraints is that many sequences have to be found where the target is close to an object of the same class, in order to make the problem tractable.
For this reason, the classes posed as an constraining object were limited to six.
In addition, the dataset contains very challenging videos to annotate, as in some sequences the constraint is positive for (many) very short sequences of frames, because the constraint object might frequently move out of proximity of the tracking target.
In other cases, some videos were easy to annotate, as the constraining object and the tracking target remained close to each other during the whole sequence.
This makes it less labour intensive to annotate, yet it could also be a downside, as this also means that the sequence might contain less information for the algorithm to learn from. 
In addition, this creates problems down the line, as the distribution of the dataset is skewed to have too many positive samples (a too high ratio of positive samples to negative samples), which is unrealistic and not desirable for performing experiments.
To tackle these issues, an effort was made to fix the distribution of samples in the dataset, which was achieved through annotating videos with relatively fewer positive samples, that required more effort to annotate. 

\section{Constrained MOT16 Dataset Details} \label{sec:appendix_ds_mot}
In this section, we specify additional details of the c-MOT16 dataset, including how the dataset was annotated, and further specifications on the details of the constraints.
The following steps describe the process of how the Constrained MOT16 dataset was created:
\begin{enumerate}
    \item Language descriptions of every tracking target in the MOT16 dataset are manually created and added.  
    \item Programmatically every tracking target $A$ is compared to every other target $B$, which acts as a constraining object for target $A$. If $B$ comes within proximity of $A$ at some point, a sequence is created where $A$ is the tracking target, and $B$ is the constraint object, and $B_l$ is its natural language description and constraint. After which for every frame of the sequence, the constraint satisfaction is set according to formula \ref{eqn:y_mot}.
\end{enumerate}

\subsection{Lingual Constraints}
In order to use the constraining object $B$ as a lingual constraint, a constraint description of the object is necessary.
As the MOT16 dataset does not include descriptions of the target objects, we added a descriptive sentence of every target object in the videos.
This description portrays the target in terms of its colours and its sub-objects.
Since most of the target objects in the MOT16 dataset are pedestrians, the annotations for these are the most descriptive, whereas other objects in the dataset, e.g. bicycles and cars (marked as occluders in the original dataset) are described by just their name and colour.
An example of a description of a pedestrian is `a person with dark shorts a green backpack and a white shirt'. 
If this is the description of the constraining object $B$, then the constraint is satisfied when $B$ walks by the tracking target $A$.
Similarly, if the characteristics of the constraining object described in the sentence are a subset of another object $C$, then the constraint is also satisfied if $C$ is nearby $A$.
An example: $B$ has description `\textit{a person with dark shorts}', then, if $C$ has description: `\textit{a person with dark shorts} a green backpack and a white shirt', then $C$ is a superset of $B$.
In this scenario, if $C$ is in proximity of $A$, the constraint is still satisfied even if $B$ is not in proximity of $A$.
We include this new edge case:
\begin{align}
\label{eqn:y_mot}
Y(A_b, B_b) &= \begin{cases}
1 & \parbox[t]{5cm}{%
    $O(B_b, A_b) \geq t$  \text{ or }  $O(C_b, A_b) \geq t$ \\
    \text{ for some } $C$ \text{ with } $B_l \subseteq C_l$,
  }
  \\
0 & \text{otherwise}, 
\end{cases}
\end{align}
in this formula, the subscript $l$, for example $C_l$, denotes the natural language description of $C$ as an object.

\section{Dataset 3: COCO for Pre-training} \label{sec:appendix_coco_pretraining}
\begin{figure}
    \centering
    \captionsetup[subfigure]{position=b,justification=centering}
    \begin{subfigure}[b]{0.22\textwidth}
     \centering
     \includegraphics[width=\textwidth]{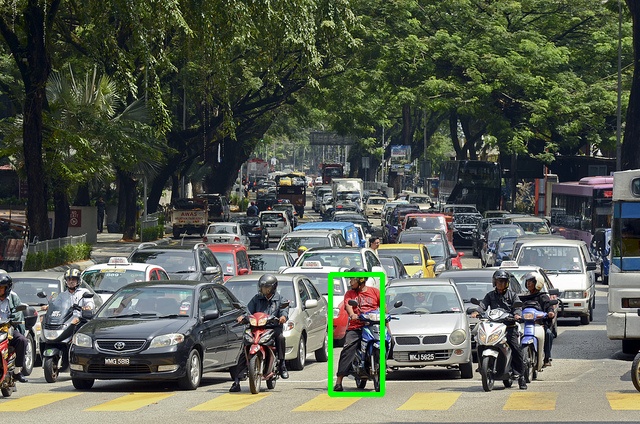}
     \caption{`besides a motorcycle'\\y: \cmark}
     \label{fig:coco_example_1}
    \end{subfigure}
    \hfill
    \begin{subfigure}[b]{0.22\textwidth}
         \centering
         \includegraphics[width=\textwidth]{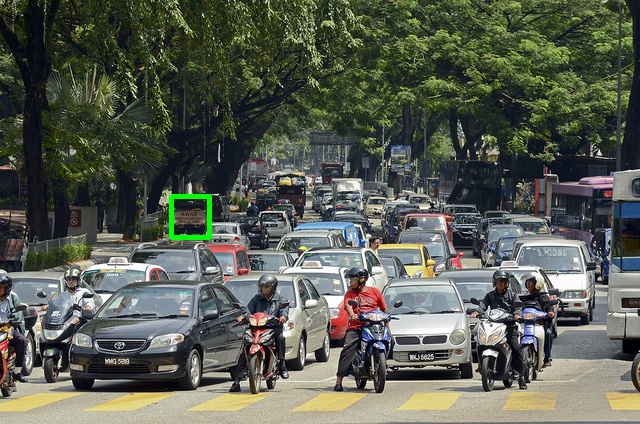}
         \caption{`close to a bird'\\y: \xmark}
         \label{fig:coco_example_2}
    \end{subfigure}
    \\
    \begin{subfigure}[t]{0.22\textwidth}
         \centering
         \includegraphics[width=\textwidth]{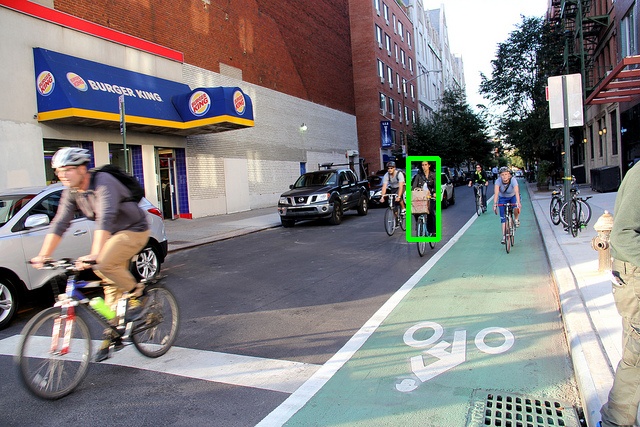}
         \caption{`besides a car'\\y: \cmark}
         \label{fig:coco_example_3}
    \end{subfigure}
    \hfill
    \begin{subfigure}[t]{0.22\textwidth}
         \centering
         \includegraphics[width=\textwidth]{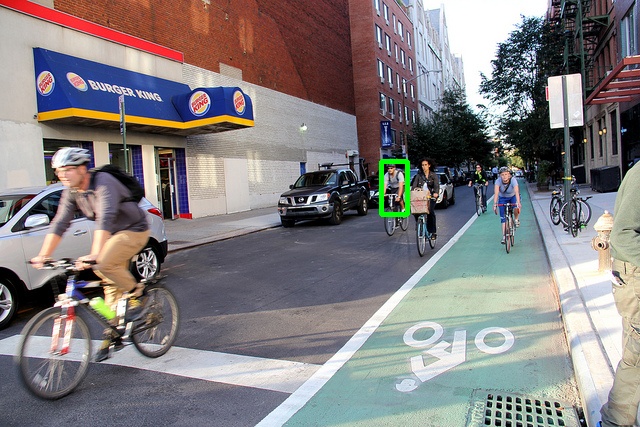}
         \caption{`alongside a backpack'\\y: \cmark}
         \label{fig:coco_example_4}
    \end{subfigure}
    \caption{\textbf{Generated pre-training data from the COCO dataset.} Shown are a few generated examples from the COCO dataset. The constraint is specified by the sentence beneath the frame and below the constraint is depicted whether or not the constraint is positive in this frame.}
    \label{fig:coco_dataset_examples}
\end{figure}

The Common Objects in Context (COCO) dataset \cite{DS:Lin2014MicrosoftCOCO} is a large scale object-detection (image) dataset, with multiple object annotations per frame.
The fact that COCO has multiple annotations per image, makes it possible to use it to pre-train our model.
The pre-training dataset is created by using one annotation as the reference and search image, and the other annotations as possible constraint examples.
Visual examples of the examples generated from the COCO dataset are shown in figure \ref{fig:coco_dataset_examples}.
This is akin to how \cite{Li2018SiamRPNEO_siamRPN++} train the SiamRPN++ model on COCO, where the search image is actually the same image as the reference image, however augmented with transformations.
This way the model learns to recognise similar images. 
The image augmentations we employ are the same augmentations used in the work of Li \etal \cite{Li2018SiamRPNEO_siamRPN++}.

\begin{table}
\centering
\caption{Examples of constraining sentences programmatically created for the COCO dataset. The left column depicts the template of the sentence, where $\{object\}$ is a placeholder for the name of the constraining object. The right column depicts an example sentence created from the template.}
\label{tab:coco-constraints}
\begin{tabular}{@{}ll@{}}
\toprule
Template                 & Example sentence               \\ \midrule
with a \{object\}        & `with a person'         \\
close to a \{object\}    & `close to a car'        \\
close by a \{object\}    & `close by a cat'       \\
adjacent to a \{object\} & `adjacent to a bicycle' \\
besides a \{object\}     & `besides a basketball'  \\
along a \{object\}       & `along a bottle'        \\ \bottomrule
\end{tabular}
\end{table}

To add the constraint annotations for the COCO dataset, we augment the data programmatically.
This is done by creating constraint data-points from objects within vicinity of the target object, where the constraint sentence is created programmatically using the name of the object.
Examples of the generated constraints are shown in table \ref{tab:coco-constraints}.
We limit the constraint objects to be an instance of one of the classes in the table \ref{tab:object-classes}, as our evaluation datasets are also limited to these classes.
An example is created when the two objects are `within vicinity', which is defined by the same overlap function $O(B_b, A_b)$ as for the Constrained MOT16 dataset. 

\section{Training}
To start with, both models are trained on the lingually constrained COCO dataset for 20 epochs, where we train on $n=\num{100000}$, randomly sampled videos per epoch.
Note that an epoch in tracking is defined differently: as frames from the same sequence (video) are often very similar and very plentiful, we do not train on all frames from all sequences each epoch.
Instead, $n$ sequences are randomly sampled each epoch.
After a sequence is sampled, the groundtruth (reference) frame is sampled from the sequence, and the search frame is sampled within a range of $100$ frames.
However, for the COCO dataset this is always the same frame, as it is an image dataset.

Consequently, the models are trained on the Constrained MOT16 dataset when evaluated on the Constrained MOT16 data, and similarly trained on the Constrained LaSOT dataset, when evaluated on that dataset.
For the Constrained MOT16 dataset, the models are optimised for up to $20$ epochs, with the same number of samples per epoch, $n=\num{100000}$.
When the model has converged, the training is stopped, which is achieved by monitoring the loss curves. 
Another way this could be achieved is with a validation set, but this was deemed too expensive as the evaluation of the validation set can take up to 12 hours. 
For the Constrained LaSOT dataset, the model is only optimised for $n=\num{5000}$ samples, as this dataset is less complex and a lot smaller in comparison, which makes the models converge much faster. 
For both models, we follow the prior work for choosing our hyper parameters as pretraining, training, and testing to evaluate a hyper parameter configuration takes up to 3 days. 
For the Constrained MOT16 dataset, we included a validation set to be able to calibrate the models, and to choose a specific decision threshold.
In the subsequent paragraphs we detail the specific hyper parameters and optimisation schemes for the models.


\subsection{\modelI{}} \label{sec:exp_model_I}
The model that uses dynamic convolutions is optimised similar as in the work of Li \etal \cite{Li2018SiamRPNEO_siamRPN++}, with a SGD optimiser.
At the start of training, the learning rate is set using a `warmup schedule' for the first 5 epochs, where it keeps a learning rate of $0.01$.
For the epochs after that, the learning rate follows a log schedule, where the starting learning rate is equal to $0.03$, and the end learning rate is equal to $\alpha = \num{5e-4}$.
Each epoch the learning rate is decreased linearly per the log space from $10^{\ln{0.03}}$ to $10^{\ln{\num{5e-4}}}$, where each epoch it decreases by a step of $\frac{10^{\ln{0.03}} - 10^{\ln{\num{5e-4}}}}{t_{total}}$, where $t_{total}$ is the total number of (non warmup) epochs.
The model is trained on a batch size of 128, which was the largest possible given the amount of VRAM that was available, as this significantly sped up training and also gave good performance.

\subsection{\modelIatt{}} \label{sec:exp_model_I_min_att}
To study the impact of the `Attention MLP', we construct a version of the model with the Attention MLP removed.
The rest of the model and training setup is left completely the same, only the output of the word embeddings processing is now directly fed to the DFG module, for reference see figure \ref{fig:dynamic-conv-model}.
In this case, the processed word embeddings are not attended by computing a weighted average with the attention weights. 
Hence, the input features of the DFG module are of size $150 \times 10$, instead of $150 \times 1$.
To accommodate this, before the linear layer in the DFG module receives the processed word embeddings, the features are concatenated as one long feature of size $(150 \times 10) \times 1$, and the size of the linear layer is changed accordingly.

\subsection{\modelII{}} \label{sec:exp_model_II}
In our experiments, the VQA models are very sensitive to the (wrong) hyperparameters.
This means that the Deep Co-Attention model will not converge at all given certain setups of learning-rates and optimizers.
In the work of Popel and Bojar~\cite{OT:Popel2018TrainingTF} further research has been done into the training of transformer models, which provided helpful knowledge for the training of the VQA models.

For the MCAN model, we follow the training setup of Yu \etal \cite{VQA:Yu2019DeepMC}.
The model was optimised using the ADAM \cite{OPTIM:Kingma2015AdamAM} optimiser, with $\beta_1 = 0.9$ and $\beta_2 = 0.98$ and a batch size of $64$. 
The model uses a base learning rate of $\alpha = \num{1e-4}$, and a warmup learning-rate factor, where this factor is set to $\gamma \in \{\frac{1}{4}, \frac{2}{4}, \frac{3}{4}\}$, for epochs 1, 2, and 3 respectively.
The factor $\gamma$ is then used as following to control the learning rate in the first 3 epochs: $ \alpha = \num{1e-4} * \gamma$.
After 10 epochs, the epoch is decreased every 2 epochs, following the following formula: $\alpha = \num{1e-4} * 0.2^{\lfloor\frac{t - 10}{2}\rfloor} $.

Furthermore, the MCAN model has several hyperparameters, such as the size of the heads and the number of MCA layers.
For these parameters we follow the parameters specified by Yu \etal \cite{VQA:Yu2019DeepMC}, as far as our computational resources allow us.
Table \ref{tab:model_II_and_ppm_params} shows a detailed overview of the hyperparameters for the MCAN model.

\subsection{\modelIIppm{}} \label{sec:exp_model_II_PPM}
To study the potential performance improvement of the PPM for lingually constrained tracking, we construct a version of \modelII{} that includes this module.
The PPM module increases the number of channels in the input, hence, the parameters of the model have to be slightly changed.
Nevertheless, to accommodate a fair comparison between the two models, they are kept as similar as possible.
The hyperparameters for the MCAN model are detailed in table \ref{tab:model_II_and_ppm_params}.  

\begin{table}[h]
\centering
\caption{\textbf{Hyperparameters for the MCAN model in \modelII{} and \modelIIppm{}.} The hyperparameters are slightly different for the attention layers, which is to accommodate the extra channels the Pyramid Pooling Module adds to the image features.}
\label{tab:model_II_and_ppm_params}
\resizebox{0.45\textwidth}{!}{%
\begin{tabular}{@{}llll@{}}
\toprule
Parameter &
  Description &
  \begin{tabular}[c]{@{}l@{}}\modelII{}\\ Value\end{tabular} &
  \begin{tabular}[c]{@{}l@{}}\modelIIppm{}\\ Value\end{tabular} \\ \midrule
$L$   & Number of MCAN layers                 & 3    & 3    \\
$d$   & Hidden layer size of the MCAN layer   & 768  & 1024 \\
$d_h$ & Size of each head in multi-head layer & 128  & 128  \\
$h$   & Number of heads                       & 6    & 8    \\
$p$   & Dropout rate                          & 0.1  & 0.1  \\
n/a &
  \begin{tabular}[c]{@{}l@{}}Feed forward size.\\ Hidden layer size of MLP in attention layer\\ Hidden layer size of Attentional Reduction\end{tabular} &
  512 &
  512 \\
n/a   & Attentional reduction output size     & 1024 & 1024 \\ \bottomrule
\end{tabular}%
}
\end{table}

\subsection{Software \& Hardware setup}
To conclude the setup of the experiments, we report the hardware and software on which the experiments were performed on.
In terms of hardware, the experiments were performed on a AMD Ryzen 5 2600X (six-core) processor, 32GB of RAM, and a Nvidia GeForce RTX 2070.
The software was built on the PyTorch \cite{OT:paszke2017automatic_pytorch}, and NumPy \cite{OT:numpy_oliphant2006guide, OT:numpy_van2011numpy} libraries.
In addition, mixed precision training was used (using the new `torch.cuda.amp' package) to further increase the available VRAM for the models.
Further credit goes to the PySOT github\footnote{https://github.com/STVIR/pysot}, and the MCAN github\footnote{https://github.com/MILVLG/mcan-vqa}, for implementations of the SiamRPN++ model and the MCAN model.

\subsection{Statistical Analysis}
Along with the quantitative analysis and qualitative analysis, it is also important to determine whether the differences in the results and models are statistically significant.
Following the work of Dietterich~\cite{OT:Dietterich1996ApproximateST}, we perform McNemar's Test, as in this case, this is the only feasible statistical test, given the computational resources required for more sophisticated tests.
As found by Dietterich~\cite{OT:Dietterich1996ApproximateST}, the test has an acceptable Type I error, i.e. the probability of detecting a difference when in fact there is no difference between the models.
This, in contrast to the popular resampled paired \textit{t} test, which has a high Type I error.
McNemar's test doesn't require as much computational resources, because it does not require retraining of the models (many times) to establish statistical significance.
Other potential statistical tests that could have been advantageous to perform, given more computational resources are: the resampled paired \textit{t} test \cite{OT:Dietterich1996ApproximateST}, the k-fold cross-validated paired \textit{t} test \cite{OT:Dietterich1996ApproximateST}, and multiple tests with different random seeds.

\section{Additional c-MOT16 Results} \label{sec:appendix_mot_results}
In this section, we report additional quantitative and qualitative results of the experiments on the c-MOT16 dataset. 

\subsection{Sequence Based Results} \label{sec:mot_seq_based_results}
\begin{figure*}
\centering
\includegraphics[width=\textwidth]{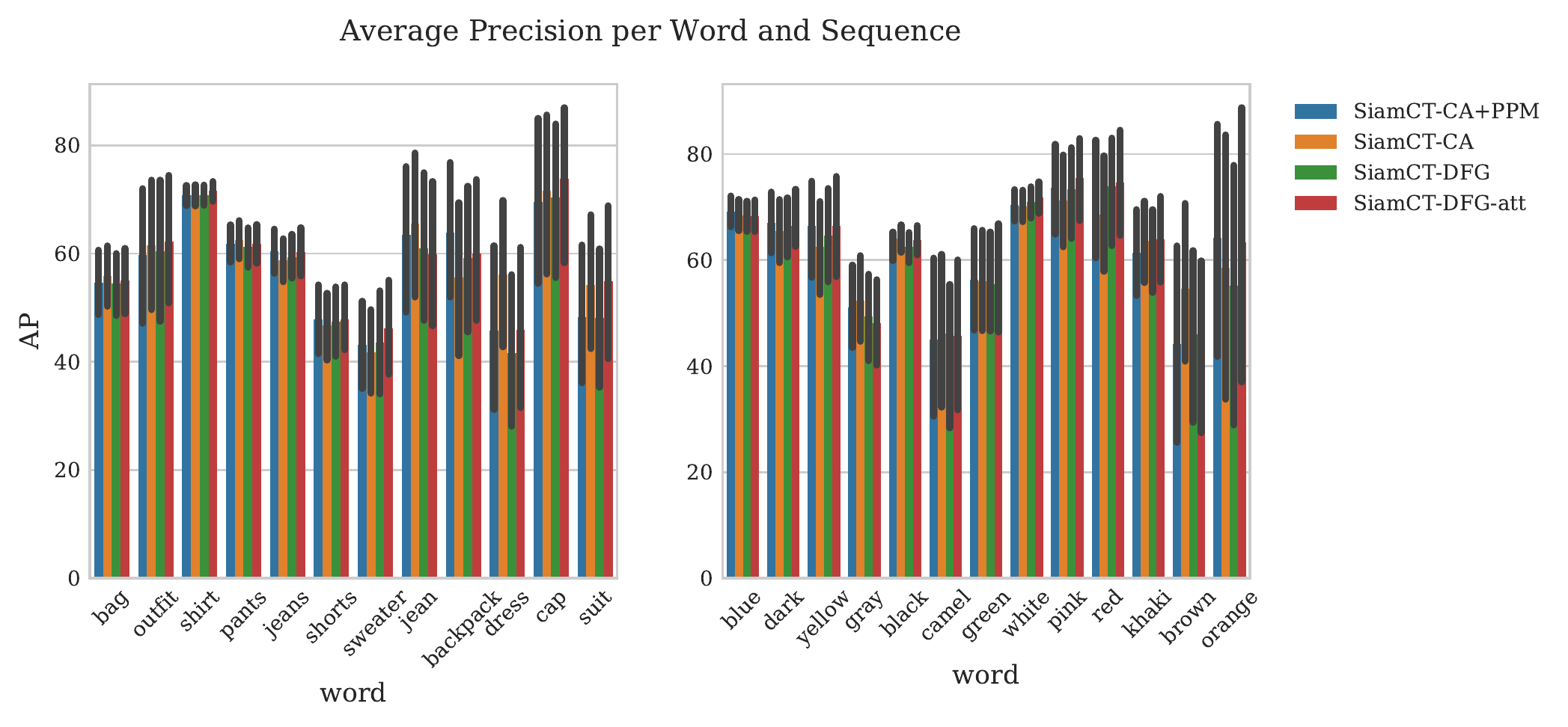}
\caption{\textbf{Results on the c-MOT16 dataset per word and sequence.} In this figure the mean of the AP per sequence is shown, and the x-axis depicts this mean for typical words in the Constrained MOT16 dataset. Error bars are plotted on top of the mean scores, calculated using a bootstrap test with a confidence interval of $0.05$.}
\label{fig:mot_video_word_performance}
\end{figure*}

We analyse the results per sequence, figure \ref{fig:mot_video_word_performance} shows similar plots as figure \ref{fig:mot_word_performance}, but now with the AP calculated per sequence after which these are averaged, instead of calculating the AP on all individual samples directly.
Moreover, since the bar plots now display the mean of a distribution of scores per sequence, error bars are plotted on top of the means.
The error bars are calculated using a bootstrap test, with a confidence-interval of $0.05$.
As visible in figure \ref{fig:mot_video_word_performance}, the variance for the results per sequence is very high, and thus it is not clear from these results whether one model is better than the other.

\subsection{Word Performance} \label{sec:mot_word_performance}
\begin{figure*}
\centering
\includegraphics[width=\textwidth]{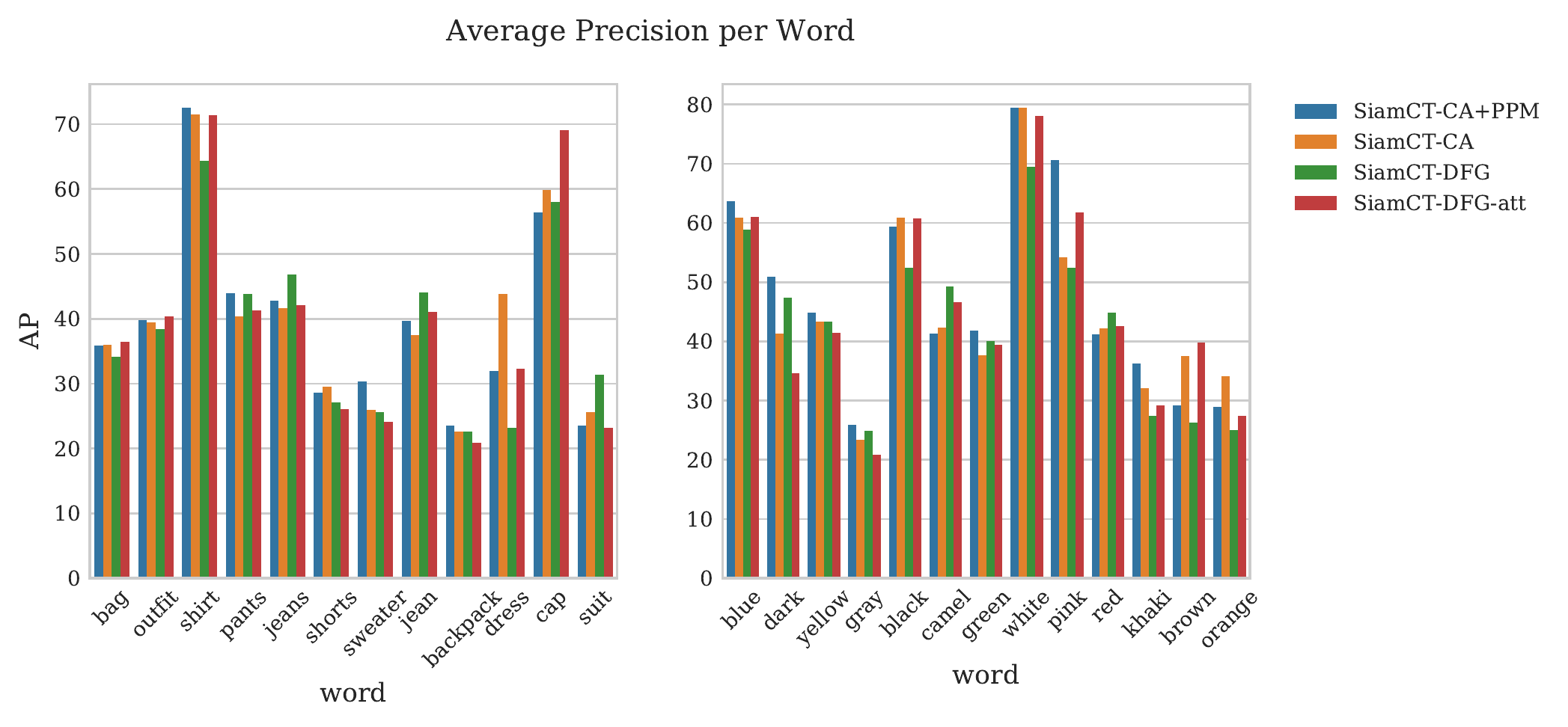}
\caption{\textbf{Word-based results on the MOT16 dataset.} In this figure the average precision is aggregated based on the words present in the constraining sentence. The LHS shows the performance for characteristic words that identify the type of object in the sentence, on the RHS the performance for colours describing these objects (in the lingual constraint) is shown.}
\label{fig:mot_word_performance}
\end{figure*}

In this section, we look at the results grouped by the words present in the constraint sentence.
Figure \ref{fig:mot_word_performance} shows the performance of the different models for the characteristic words in the dataset.
The plot on the left hand side shows the average precision for the different constraining objects present in the dataset, and the right hand side shows the average precision for the different colours present in the constraints.
The results of these graphs are obtained by grouping the predictions based on whether the word in question is present in the constraint sentence.
These words are important, as these words describe the constraining object, which the classifier has to visually find in the search frame to determine whether the constraint is in the frame or not.
In this case, we only show the results for the Average Precision metric, because it is threshold insensitive, and thus does not create misleading results.
As shown in the plots, the models score very close to each other on most words, and the overall AP is relatively low for many of the words, especially for the objects.
The words that have better scores are the ones more frequently present in the dataset overall; namely, `shirt', `cap', `blue', `black', and `white'.
Other than that, the differences between the models per word are all very small, except some outliers that every model seems to have one or two of.

\subsection{McNemar's Test} \label{sec:mot_mcnemars_stat_test}
Lastly, we performed McNemar's test to evaluate the significance of the differences between the predictions of each model.
Table \ref{tab:mcnemars_mot} shows the probability that the null-hypothesis is correct for all pairs of models (the null-hypothesis is that the two models are the same).
The two models are significantly different if the probability $p$ is below $p < 0.05$, which is clearly the case for all tests.
This result evidently indicates that all the classifiers are significantly different.

\begin{table}[h]
\centering
\caption{\textbf{c-MOT16 McNemar's Test Results.} Significance testing of differences between models. Entries in the cells display the null-hypothesis probability. In other words the probability that the models are not different. Diagonal entries and mirrored entries are marked n/a, short for non-applicable.}
\label{tab:mcnemars_mot}
\resizebox{0.45\textwidth}{!}{%
\begin{tabular}{@{}lllll@{}}
\toprule
               & \modelI{} & \modelIatt{} & \modelII{} & \modelIIppm{} \\ \midrule
\modelI{}        & n/a     & n/a           & n/a      & n/a            \\
\modelIatt{}  & 0.0     & n/a           & n/a      & n/a            \\
\modelII{}       & 0.0     & 0.0           & n/a      & n/a            \\
\modelIIppm{} & 0.0     & 0.0           & 1.41e-14 & n/a            \\ \bottomrule
\end{tabular}%
}
\end{table}

\subsection{Attention Maps} \label{sec:mot_attention_maps}
\begin{figure*}[bt]
    \centering
\makebox[\textwidth][c]{
\begin{tabular}{c|c}
\begin{subfigure}[t]{0.5\textwidth}
    \includegraphics[width=\textwidth, height=5cm]
    {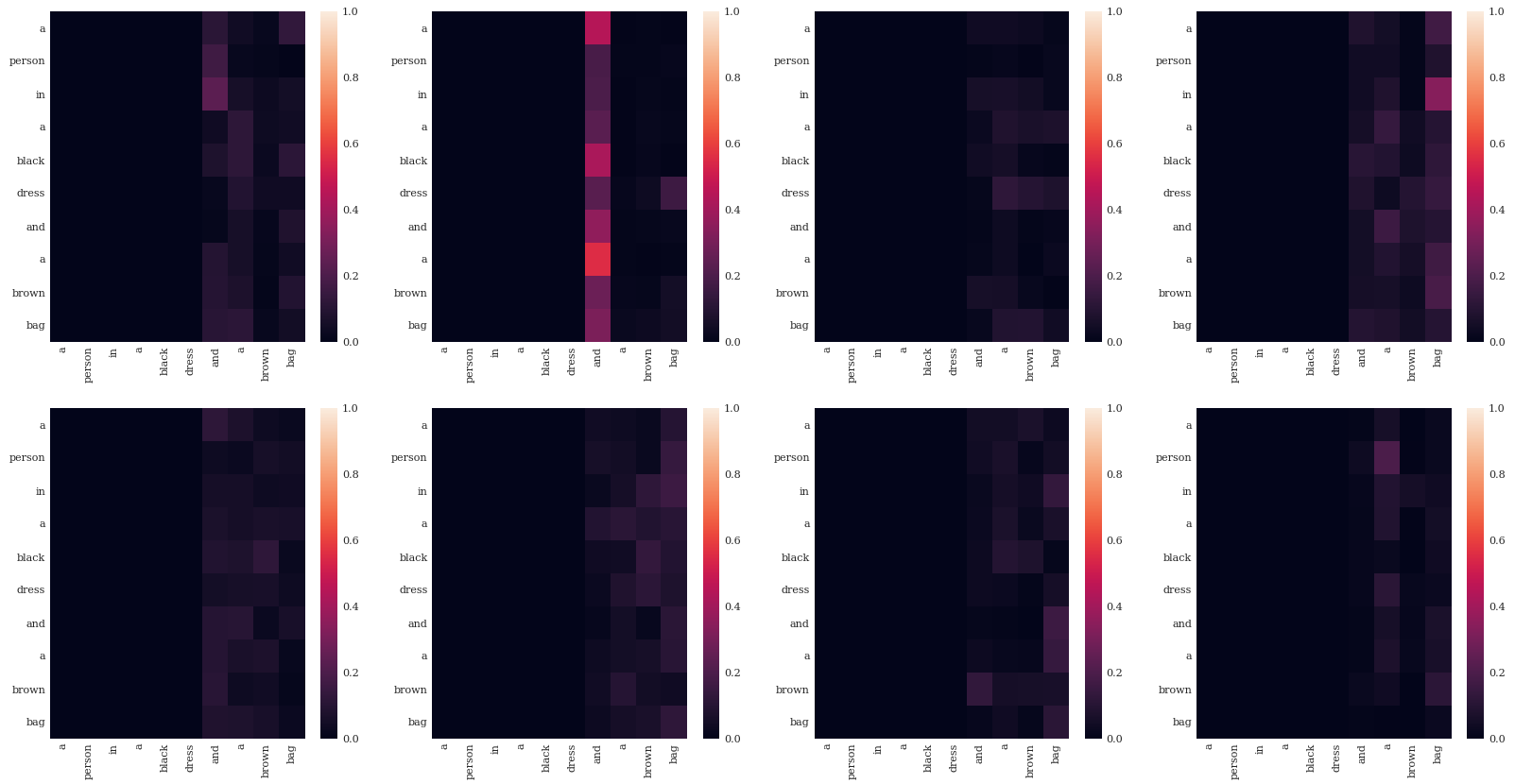}
    \caption{SA-3}
\end{subfigure}\hfill
&
\begin{subfigure}[t]{0.5\textwidth}
    \includegraphics[width=\textwidth, height=5cm]{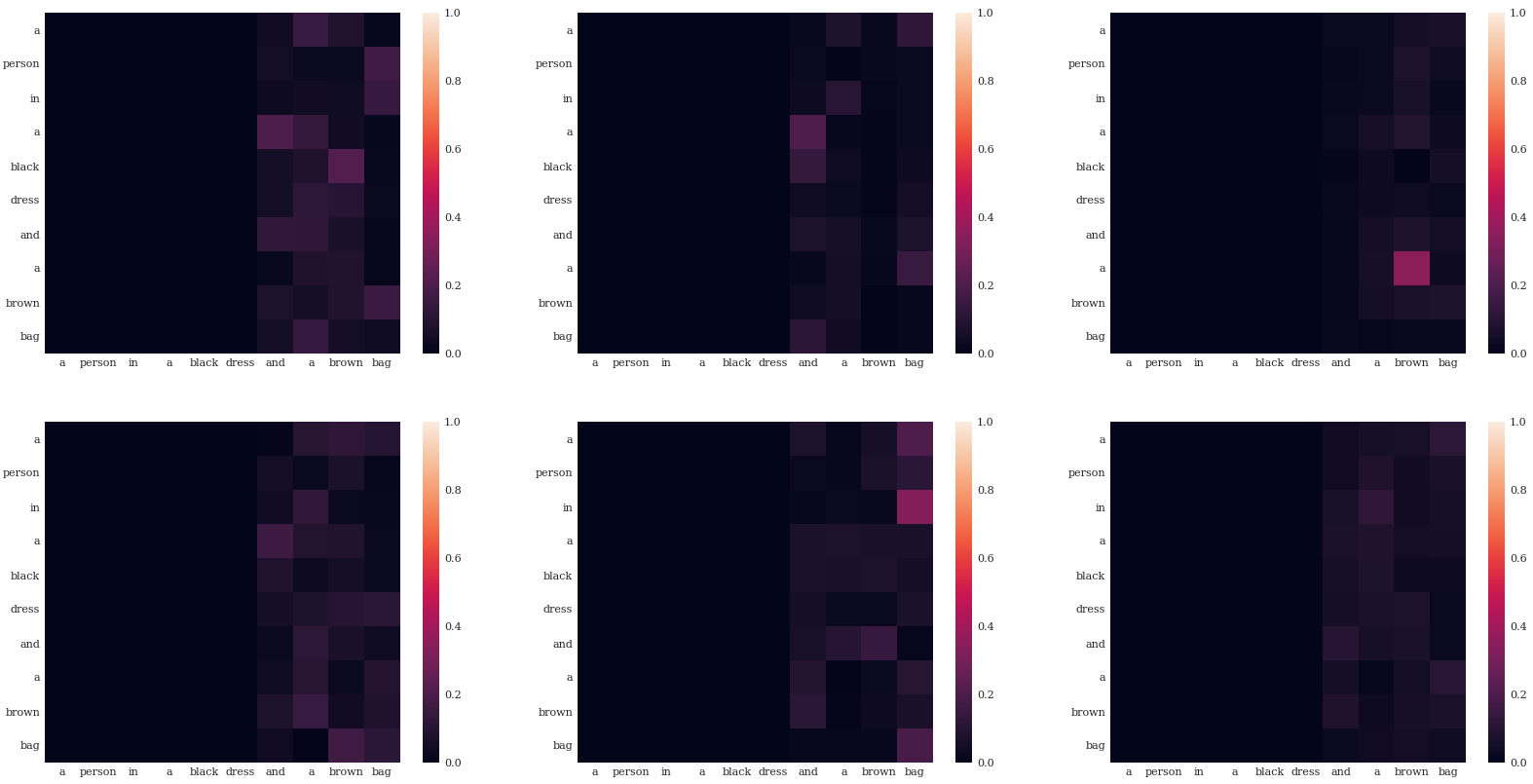}
    \caption{SA-3}
\end{subfigure}
\end{tabular}
}%
\\
\makebox[\textwidth][c]{
\begin{tabular}{c|c}
\begin{subfigure}[t]{0.5\textwidth}
     \includegraphics[width=\textwidth, height=5cm]
     {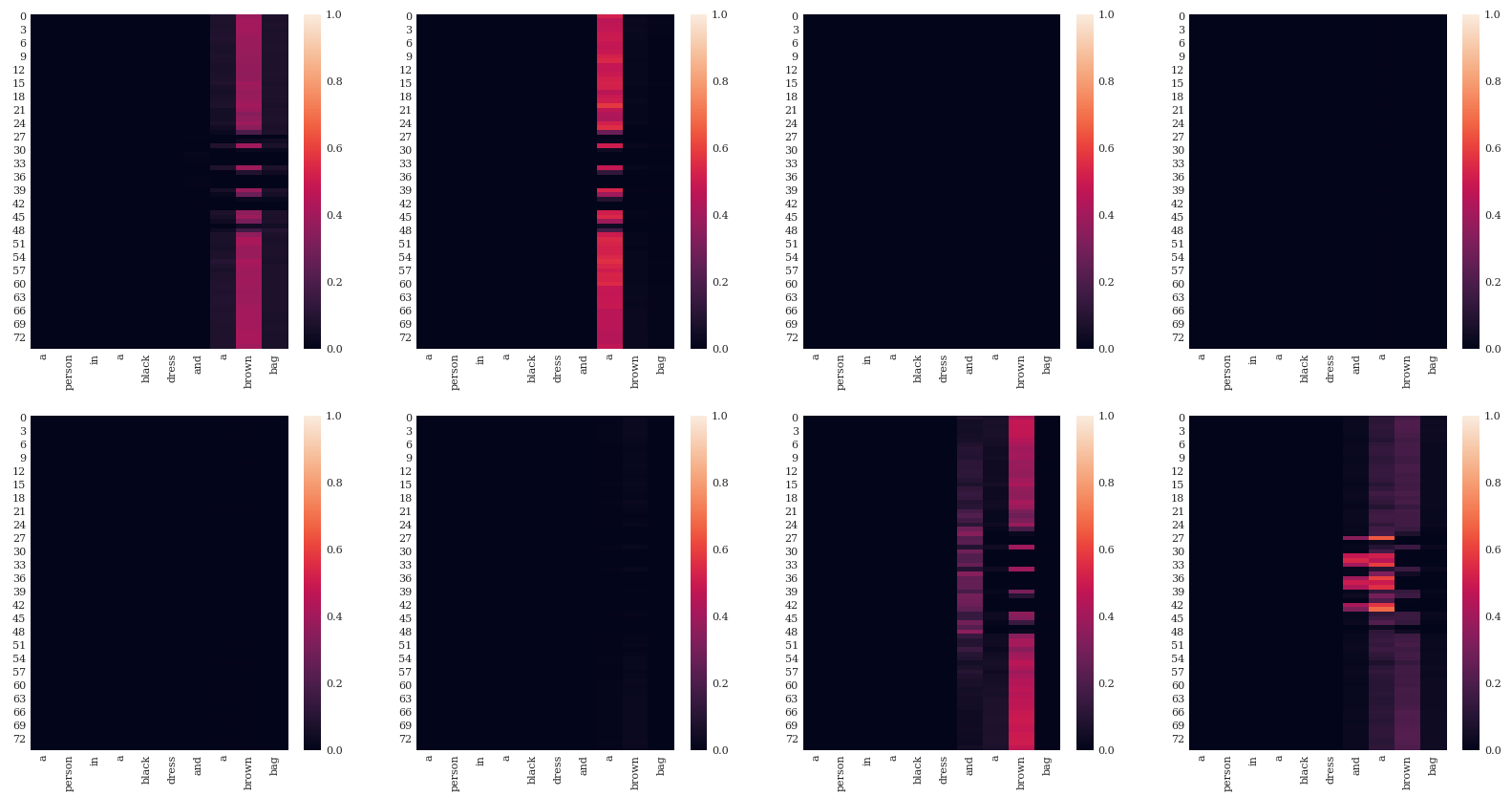}
     \caption{SGA-3\\ \modelIIppm{}}
\end{subfigure}\hfill
&
\begin{subfigure}[t]{0.5\textwidth}
    \includegraphics[width=\textwidth, height=5cm]{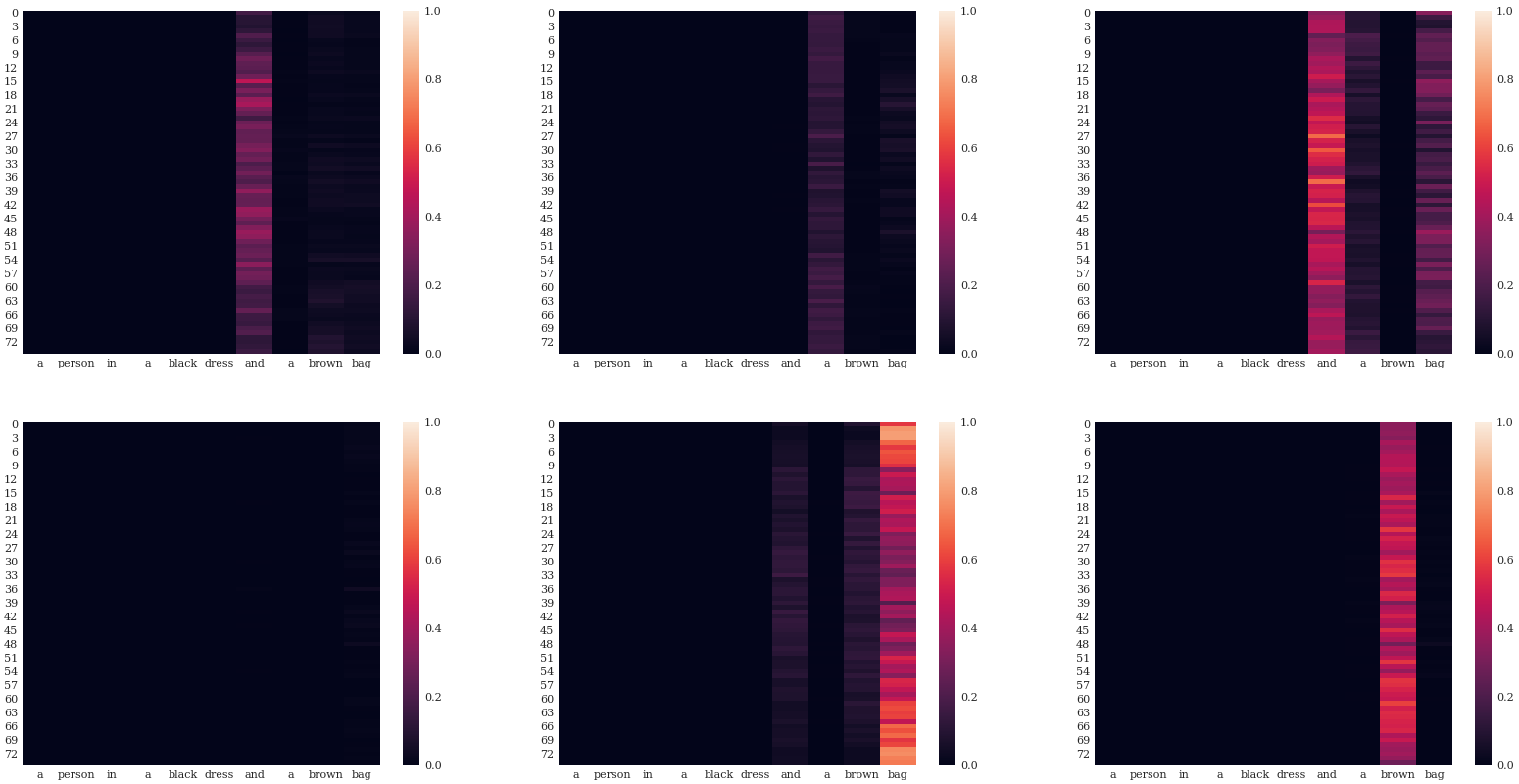}
    \caption{SGA-3\\ \modelII{}}
\end{subfigure}
\end{tabular}
}%
    \caption{\textbf{Attention Maps on the Constrained MOT16 dataset.} Attention maps are shown for the image in the first row in figure \ref{fig:mot_visual_predictions}, with the lingual constraint ``a person in a black dress and a brown bag''. The first 2 rows depict the multi-head attention weights (one plot per head) from layers SA-3 (the last self-attention layer), which self-attends the lingual constraint. The last 2 rows display the multi-head SGA-3 (self-guided attention) layers, which attend the image based on the constraint. Best viewed in colour and zoomed in.}
    \label{fig:mot_attention_maps}
\end{figure*}

In this section, we look more in depth at the attention layers of \modelII{} and \modelIIppm{}.
Figure \ref{fig:mot_attention_maps} shows the attention maps for the first image in figure \ref{fig:mot_visual_predictions}.
Displayed are the attention maps from the last lingual constraint self-attention layer (SA-3), and the last guided attention layer based on the sentence (SGA-3).

What stands out for the SA-3 layer, is that there is no self-attention on the first five words in the constraint, as all attention is in the last columns.
A stark difference between the models in the SA-3 layer is that \modelIIppm{} has one attention head that heavily focuses on the word `and', which is not the case in any of the attention heads of \modelII{}.
Overall, the majority of the attention is on the words `and a brown bag', and the other important words in the sentence, i.e., `black dress' are mostly unattended.

In the SGA-3 layers, the image is attended based upon the lingual constraint.
First, there are several attention heads that have mostly low attention weights everywhere, this could indicate that the $8$ or $6$ number of heads, are not entirely necessary.
If this is the case, the number of heads could be reduced to increase the computational efficiency of the model.
Second, we see that each head seems to attend to almost entirely one word, and attend the image based upon that.
Having said that, it is clear that the words cannot directly be related to one image patch, as `brown' and especially `bag' are not present everywhere in the image.

\section{Additional Constrained LaSOT Results} \label{sec:appendix_lasot_results}
In this section, we report additional quantitative and qualitative results of the experiments on the c-LaSOT dataset. 

\subsection{McNemar's Test}
If we now turn to McNemar's test in table \ref{tab:mcnemars_lasot}, we evaluate whether the models are statistically different.
In table \ref{tab:mcnemars_lasot} the probability of the null-hypothesis being true are depicted.
The results show a similar trend to the results for the MOT dataset, as the probabilities indicate that the differences between the models are significant.

\begin{table}[h]
\centering
\caption{\textbf{McNemar's Test Results.} Significance testing of difference between models, entries in the cells display the null-hypothesis probability, namely that the models are not different. Diagonal entries and mirrored entries are marked n/a, short for non-applicable.}
\label{tab:mcnemars_lasot}
\resizebox{0.45\textwidth}{!}{%
\begin{tabular}{@{}lllll@{}}
\toprule
               & \modelI{}   & \modelIatt{} & \modelII{}  & \modelIIppm{} \\ \midrule
\modelI{}        & n/a       & n/a           & n/a       & n/a            \\
\modelIatt{}  & 1.21e-261 & n/a           & n/a       & n/a            \\
\modelII{}       & 0.0       & 1.08e-99      & n/a       & n/a            \\
\modelIIppm{} & 0.0       & 0.0           & 3.34e-125 & n/a            \\ \bottomrule
\end{tabular}%
}
\end{table}

\subsection{Word Performance \& Sequence Based Results}
\begin{figure*}[ht]
\centering
\includegraphics[width=\textwidth]{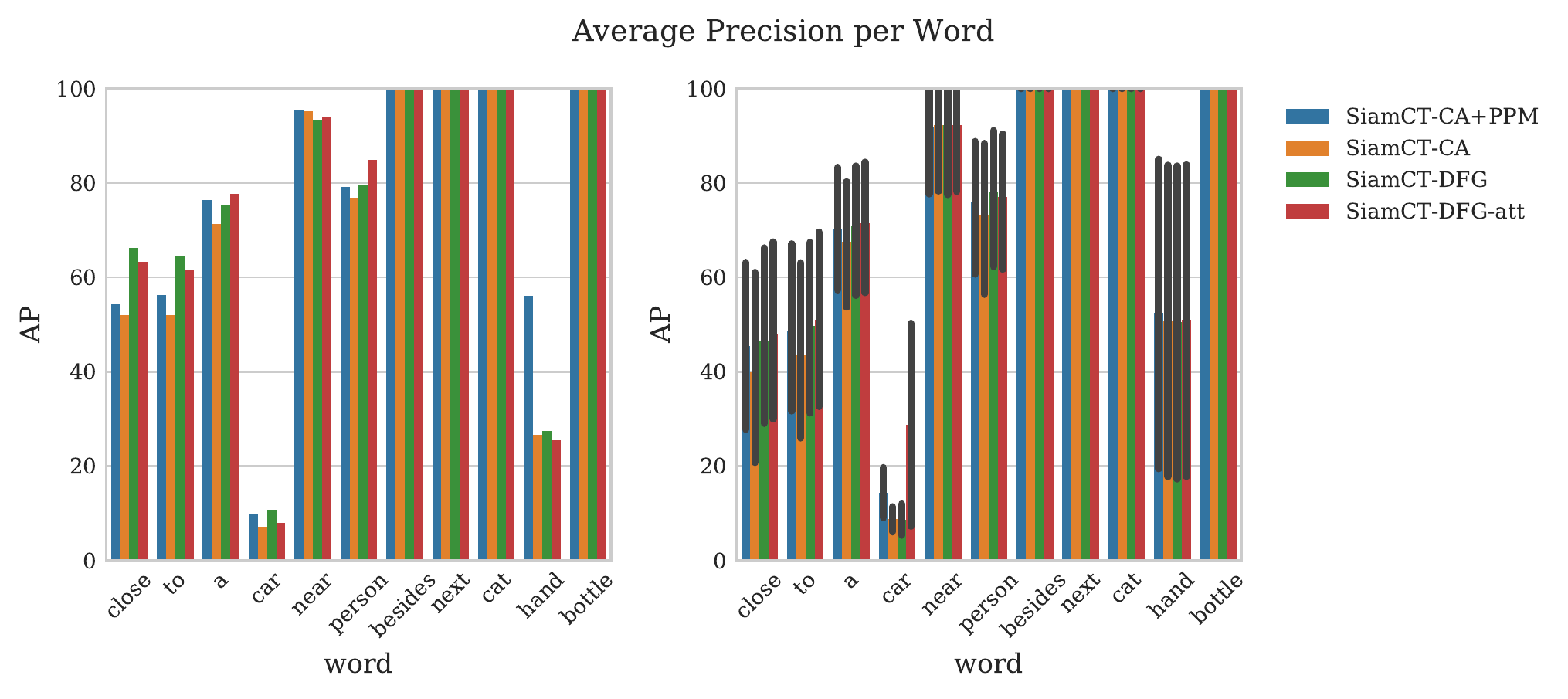}
\caption{\textbf{Results on the LaSOT dataset per word.} This figure shows the AP calculated for typical words in the lingual constraints of the Constrained LaSOT dataset. The LHS shows the overall average precision. The RHS shows the mean average precision for all sequences.}
\label{fig:lasot_word_performance}
\end{figure*}

In this section, we analyse the results per word in the lingual constraints, as well as doing a sequence-based analysis on these words.
Figure \ref{fig:lasot_word_performance} shows the performance on the (small set) of different words in the LaSOT dataset, where the LHS shows the AP per word for all sequences together (non-averaged), and the RHS shows the average performance per word on the sequences.
Similarly to figure \ref{fig:mot_video_word_performance} for the MOT dataset, the error bars are plotted on the sequence based averages to show the confidence interval of the mean.

From the words plotted, there are some `general' words, that may occur in many of the constrained tracks, like `close', `to', `besides' and `near', these essentially show the general performance on the dataset.
The other words are more specific per sample (and important for the classification), like `person', `cat', `hand', `car' and `bottle'.
On the LHS, we denote that the difference in AP is very small for most words, only for the word `hand' a larger improvement is present for \modelIIppm.
Additionally, the predictions for the word `hand' and `car' are clear outliers in terms of performance, compared to the rest of the words.
On the RHS we see a similar trend for most words.
The mean AP performance is very close for all models. 
Only \modelIatt has on average a better performance for the word `car'. Even so, the broad confidence interval shows we cannot be confident when comparing the models with these statistics.

\subsection{Attention Maps} \label{sec:lasot_attention_maps}
\begin{figure*}[hbt]
    \centering
\makebox[0.98\textwidth][c]{
\begin{tabular}{c|c}
\begin{subfigure}[t]{0.48\textwidth}
    \includegraphics[width=\textwidth, height=5cm]{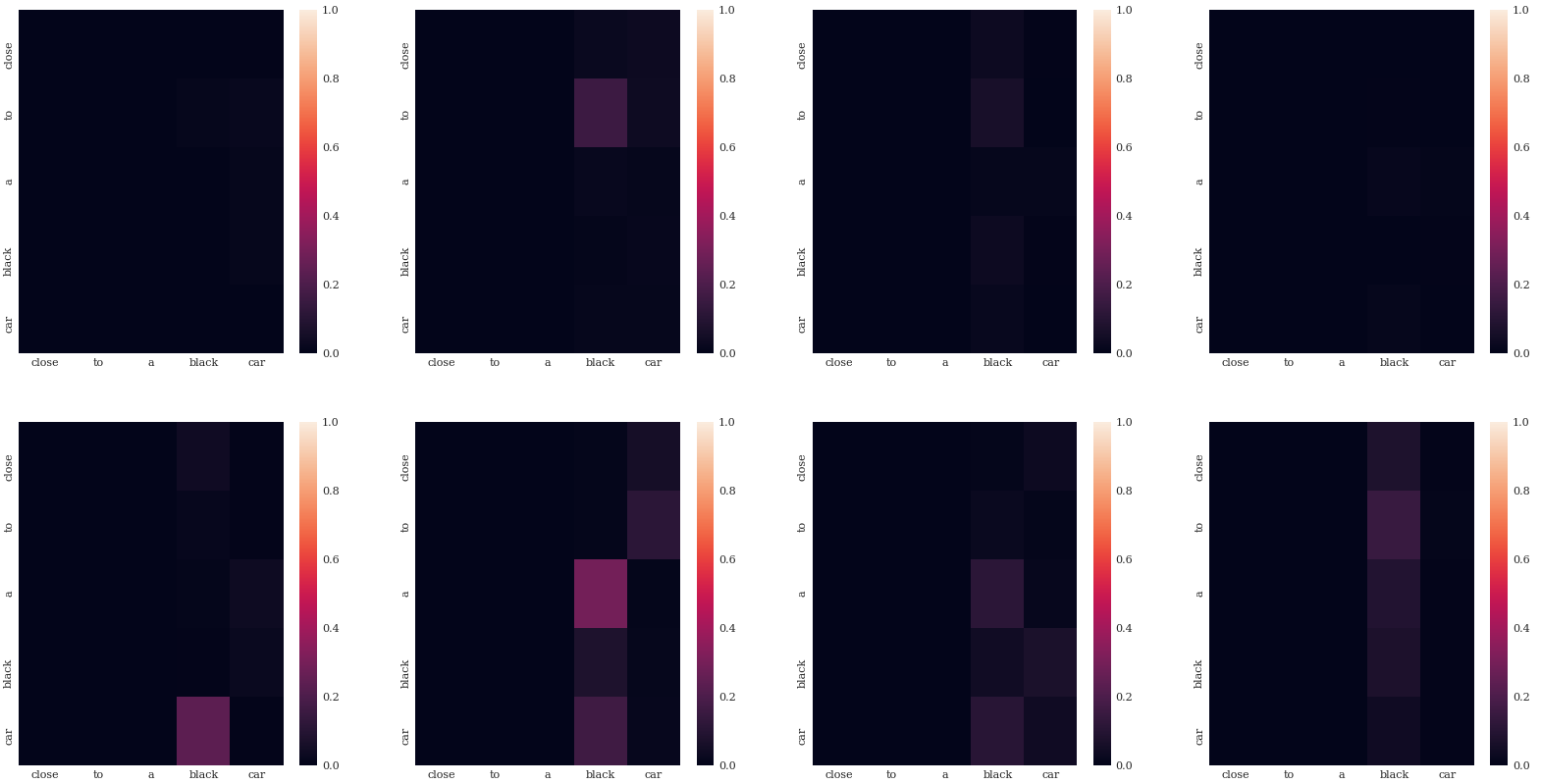}
    \caption{SA-3}
\end{subfigure}\hfill
&
\begin{subfigure}[t]{0.48\textwidth}
    \includegraphics[width=\textwidth, height=5cm]{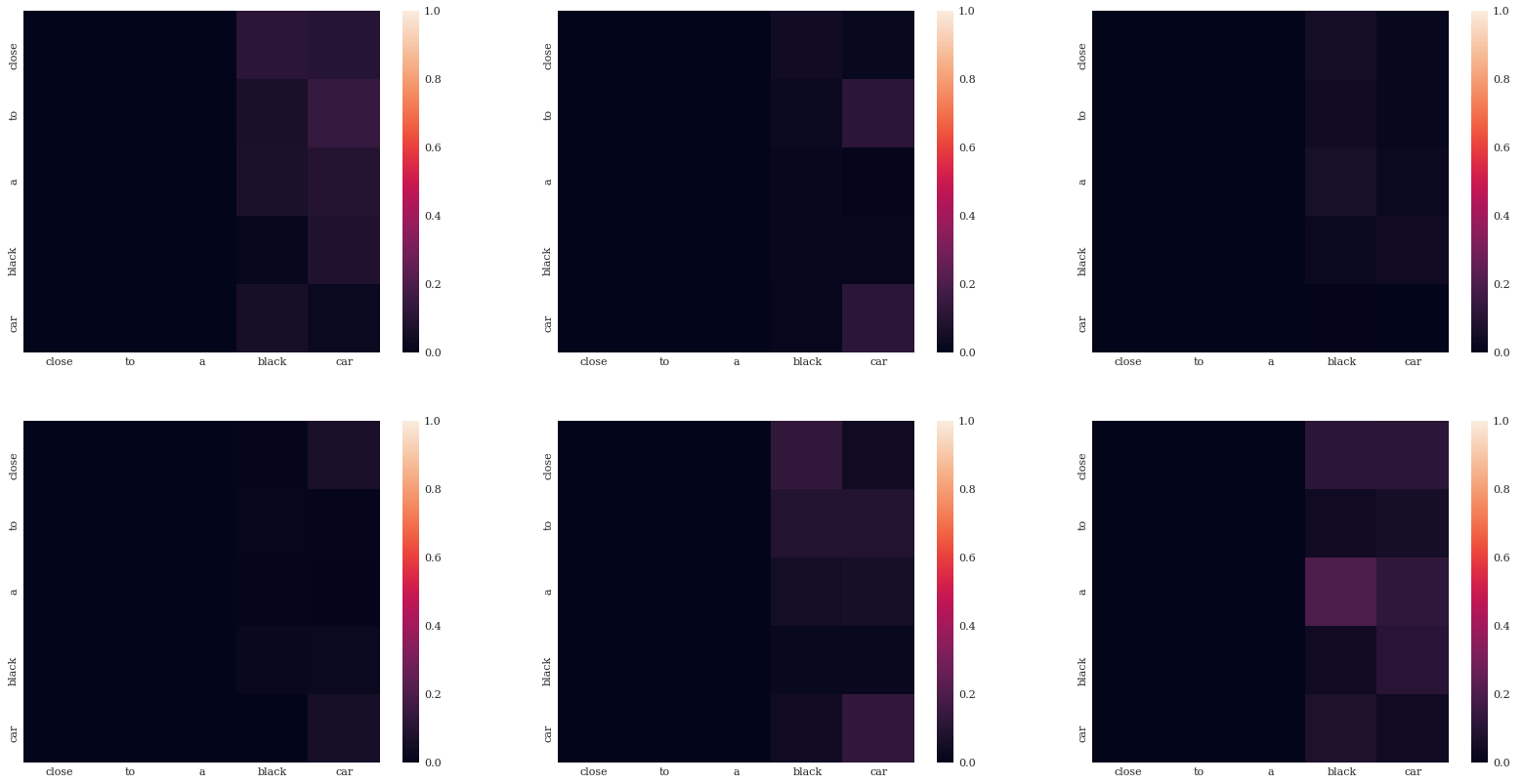}
    \caption{SA-3}
\end{subfigure}
\end{tabular}
}%
\\
\makebox[0.98\textwidth][c]{
\begin{tabular}{c|c}
\begin{subfigure}[t]{0.48\textwidth}
     \includegraphics[width=\textwidth, height=5cm]{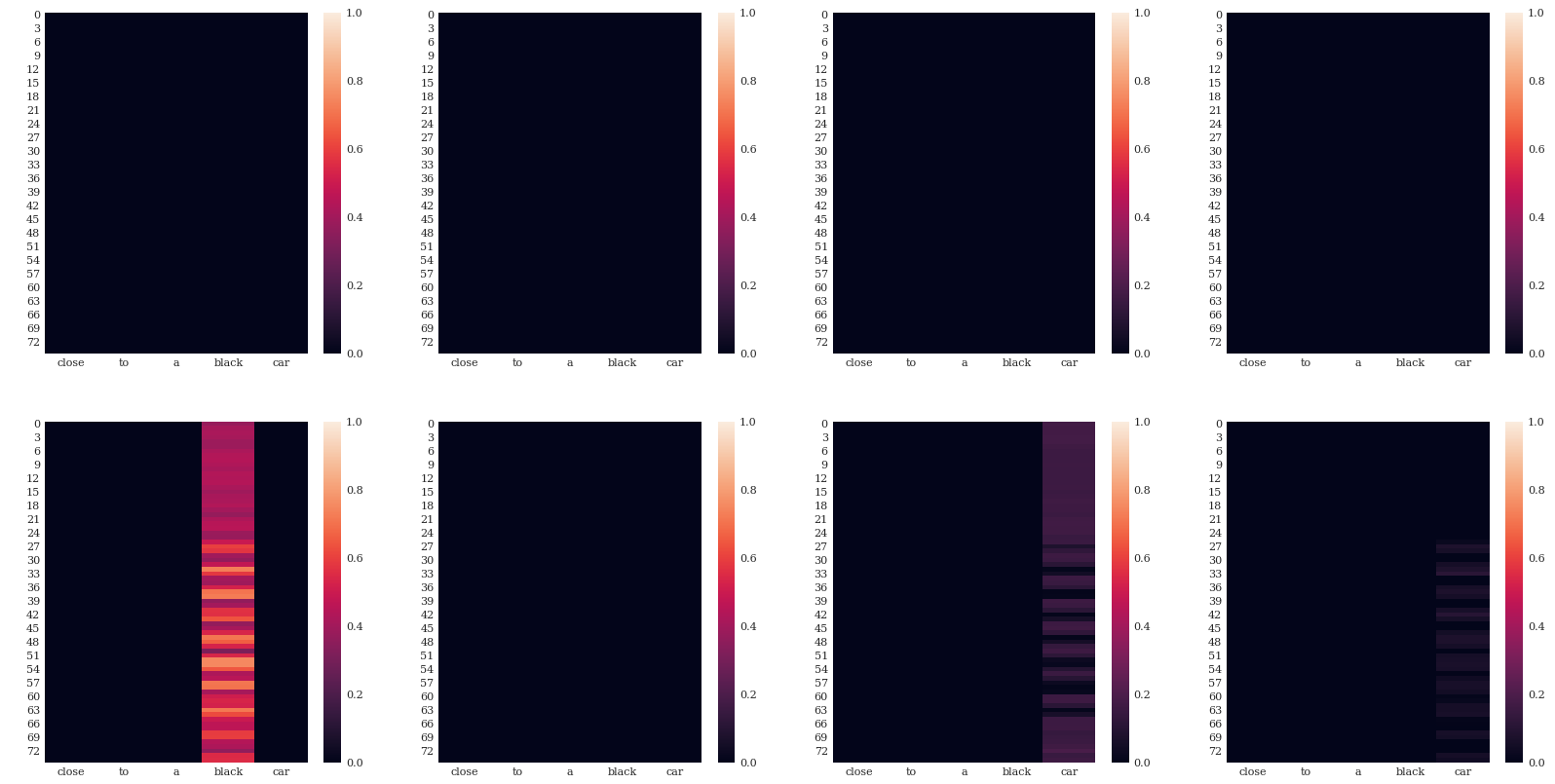}
     \caption{SGA-3\\ \modelIIppm{}}
\end{subfigure}\hfill
&
\begin{subfigure}[t]{0.48\textwidth}
    \includegraphics[width=\textwidth, height=5cm]{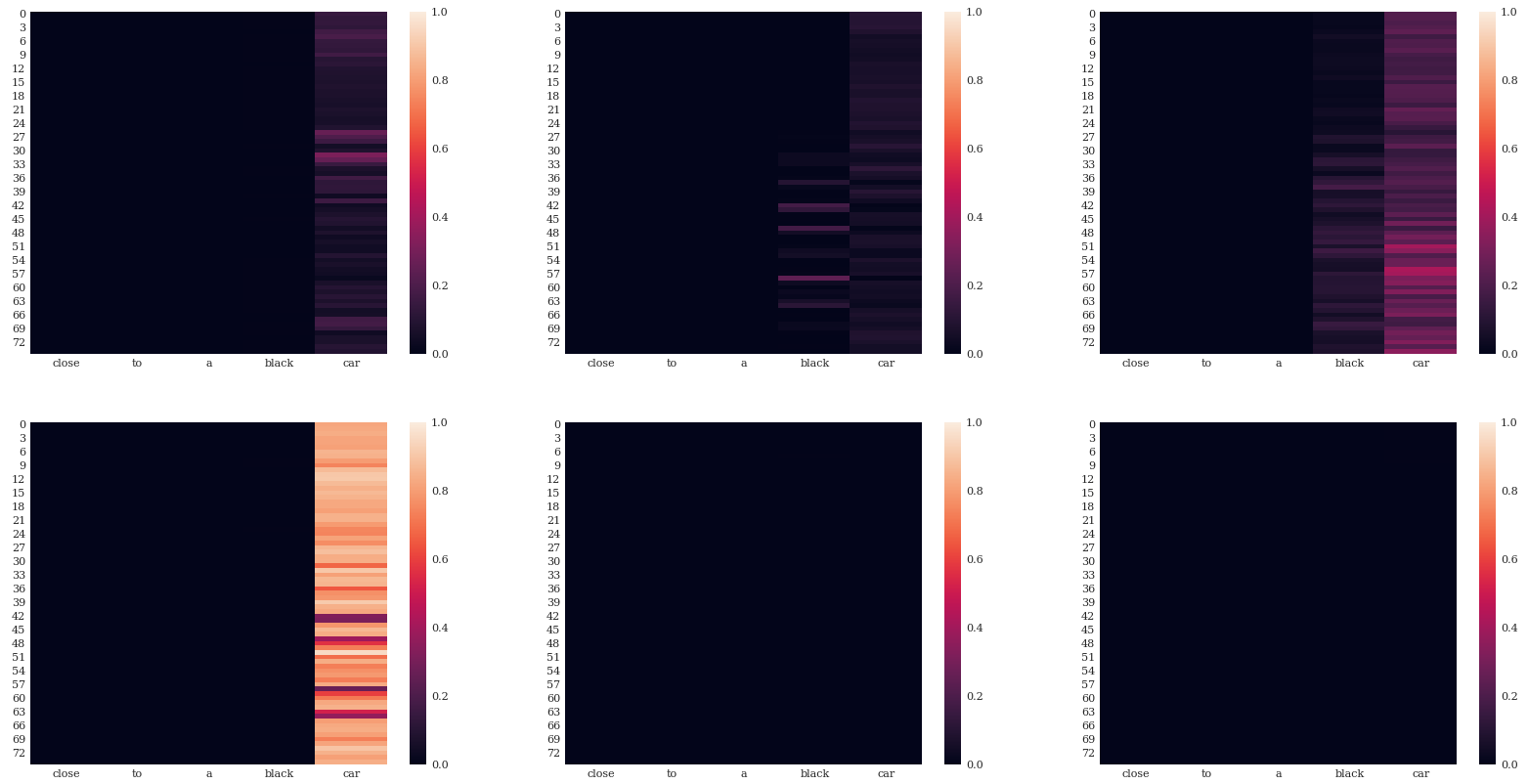}
    \caption{SGA-3\\ \modelII{}}
\end{subfigure}
\end{tabular}
}%
    \caption{\textbf{Self-attention maps \& self-guided attention maps on the Constrained LaSOT dataset.} Attention maps are shown for the image in the first row in figure \ref{fig:mot_visual_predictions}, with the lingual constraint ``close to a black car''. The LHS shows \modelIIppm{} attention maps (in the first 4 columns) and the RHS shows the maps for \modelII{} (last 3 columns). The first 2 rows depict the multi-head attention weights (one plot per head) from layers SA-3 (the last self-attention layer), which self-attends the lingual constraint. The last 2 rows display the multi-head SGA-3 (self-guided attention) layers, which attend the image based on the constraint. Since \modelIIppm{} has a slightly different architecture, it has 8 heads instead of 6, which explains that the LHS has 8 plots per layer, in contrast to the 6 of \modelII{}.}
    \label{fig:lasot_attention_maps}
\end{figure*}

After inspecting the visual predictions in the last section, we now look more closely at the attention maps of the first example (`close to a black car') in figure \ref{fig:mot_visual_predictions}. 
Figure \ref{fig:lasot_attention_maps} displays the attention maps for the last self-attention (SA-3) layer that attends the lingual constraint.
Additionally, it displays the last self-guided attention layer (SGA-3) that attends the image based on the lingual constraint.

For the SA-3 layer, the first stark contrast between the models is that \modelII{} has concentrated attention on the word `car', while the variant with the Pyramid Pooling Module has more spread out self-attention in layer SA-3.
Similarly, the same pattern shows itself for the guided-attention layers, where the \modelII{} has  almost all of its heads only attending the word `car', in contrast, \modelIIppm{} also attends the word `black' in its fifth head.
In addition, we see a similar trend as in the MOT dataset, where 7 out of the 14 attention layers (between the models) do not attend any visual patch strongly.

\end{document}